\documentclass{article}

\usepackage[preprint]{neurips_2026}
\usepackage{enumitem}
\usepackage{graphicx}
\usepackage{wrapfig} 
\usepackage{subcaption} 
\usepackage{caption}
\usepackage{amsmath,amssymb}
\usepackage{multirow} 
\usepackage{float} 
\usepackage[utf8]{inputenc} 
\usepackage[T1]{fontenc}    
\usepackage{hyperref}       
\usepackage{url}            
\usepackage{booktabs}       
\usepackage{amsfonts}      
\usepackage{nicefrac}     
\usepackage{microtype}     
\usepackage{xcolor}         
\usepackage{booktabs} 
\usepackage{bm}    
\usepackage{algorithm}
\usepackage{algorithmic}

\usepackage{amsthm}
\newtheorem{proposition}{Proposition}

\newcommand{\figrefsub}[2]{\hyperref[#1]{\ref{#1}\hspace{1.5pt}(#2)}}

\newcommand{\sysname}{EMTC}

\title{Episodic Memory Temporal Consistency for Cooperative Multi-Agent Reinforcement Learning}
\usepackage{titlesec}
\usepackage{amssymb}
\titlespacing{\section}{0pt}{0.2\baselineskip}{0.2\baselineskip}

\titlespacing{\subsection}{0pt}{0.1\baselineskip}{0.1\baselineskip}
\author{%
  Zicheng Zhao$^{1}$, Yu Lan$^{1}$, Chengzhengxu Li$^{1}$, \\
  \textbf{Zhaohan Zhang$^{2}$, Xiaoming Liu$^{1,*}$} \\
  \small $^{1}$Xi'an Jiaotong University \quad $^{2}$Queen Mary University of London \\
  \small \texttt{\{zzc0109, ylan2020, czx.li, xm.liu\}@xjtu.edu.cn} \\
  \small \texttt{\{zhaohan.zhang\}@qmul.ac.uk} \\
  \small $^*$Corresponding author
}
\begin{document}

\maketitle

\begin{abstract}

Cooperative Multi-Agent Reinforcement Learning (MARL) frequently suffers from severe reward sparsity and exploration bottlenecks. While episodic memory mechanisms mitigate these issues by reusing high-return trajectories, they often trap agents in local optima due to unconstrained incentive distribution and semantic representation collapse. To address this, we propose Episodic Memory Temporal Consistency (EMTC), a framework that robustly constructs and selectively leverages historical experiences. EMTC introduces two synergistic components: (1) a Temporally Consistent Semantic Embedder that integrates contrastive learning with time-conditioned state reconstruction, preventing representation collapse and enabling precise memory retrieval; and (2) a Temporal Consistency Gating Mechanism that dynamically modulates episodic incentives based on temporal consistency error. This adaptive gate filters misleading signals from pseudo-successful trajectories, effectively mitigating Q-value overestimation. We provide theoretical guarantees, establishing a strict error bound that directly links the observable temporal consistency error to the underlying trajectory optimality and representation quality. Extensive evaluations on the SMAC and GRF benchmarks demonstrate that EMTC consistently outperforms state-of-the-art baselines. Notably, compared to the strongest episodic baseline, EMTC achieves absolute win-rate improvements of up to 24\% in super-hard SMAC scenarios and an average improvement of 28\% across GRF tasks.

\end{abstract}

\section{Introduction}
\label{intro}
Recently, cooperative multi-agent systems have demonstrated remarkable success in tackling complex problems across a wide spectrum of domains, including robotics (\cite{chen2025multi}), cooperative autonomous driving (\cite{hao2025research,zhou2022multi}), and communication networks (\cite{hong2024multi}). Nevertheless, deploying these systems in real-world applications still faces formidable challenges, mainly stemming from inherent partial observability and highly complex coordination dynamics among interacting agents.

To tackle these challenges, Centralized Training with Decentralized Execution (CTDE) (\cite{oliehoek2008optimal,oliehoek2016concise,gupta2017cooperative}) and value factorization algorithms (e.g., QMIX (\cite{rashid2020monotonic}), DMIX (\cite{sun2021dfac}), and QPLEX (\cite{wang2020qplex}) have emerged as the mainstream paradigm in cooperative MARL (\cite{samvelyan2019starcraft}). Despite their success, these value-based methods fundamentally rely on dense step-wise rewards, suffering from profound exploration bottlenecks and suboptimal entrapment in environments with severe reward sparsity and vast joint action spaces. To mitigate this, episodic memory mechanisms have been integrated into MARL (\cite{blundell2016model,lengyel2007hippocampal,lin2018episodic,pritzel2017neural}). By caching historical high-return trajectories to generate intrinsic incentives, recent frameworks like EMC (\cite{zheng2021episodic}) and EMU (\cite{na2024efficient}) break initial exploration deadlocks and significantly accelerate convergence. In a typical episodic memory workflow, a parametric encoder first maps visited states into a compact latent space, where high-return trajectories are stored as key-value pairs. During subsequent exploration, agents retrieve the maximum historical return associated with the current state's nearest neighbor in this latent space. This retrieved value is then used to construct an intrinsic incentive that guides the policy toward previously successful regions.
A comprehensive discussion of related work is provided in Appendix~\ref{app:related}.

\begin{wrapfigure}{r}{0.6\textwidth} 
\vspace{-10pt}
    \centering
    \includegraphics[width=0.60\textwidth]{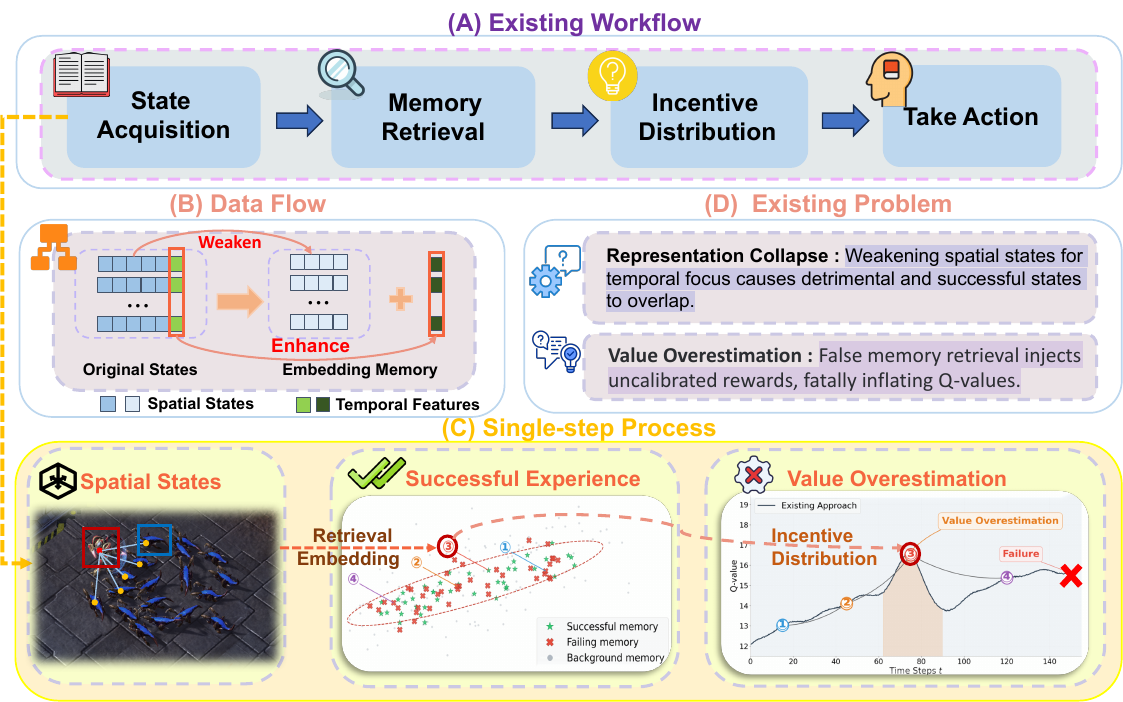} 
    \caption{\textbf{The coupled bottlenecks of conventional Episodic Memory in MARL.} \textbf{(A)} The standard episodic workflow. \textbf{(B)} Biased data flow weakens spatial states in favor of temporal features. \textbf{(C)} A detrimental spatial state is falsely embedded into a successful cluster (probe \textcircled{3}), which subsequently triggers an uncalibrated reward injection and a fatal Q-value spike (Value Overestimation). \textbf{(D)} Summary of the existing problems.}
    \label{fig:workflow}
    \vspace{-8pt}
\end{wrapfigure}

However, this initial benefit proves transient, as conventional memory workflows (Figure~\ref{fig:workflow}(A)) suffer from a coupled architectural flaw: \textbf{imbalanced feature prioritization during state embedding}. By aggressively modeling temporal sequences (Figure~\ref{fig:workflow}(B)), standard encoders inadvertently suppress discriminative spatial semantics\footnote {Fundamentally, \textit{spatial semantics} denotes an agent's structural awareness of its environment. While instantiated as explicit coordinate arrays in our tasks, this concept directly aligns with the visual/3D spatial representations in Embodied AI.}, triggering a detrimental cascade of failures summarized as two coupled bottlenecks (Figure~\ref{fig:workflow}(D)): 
\textbf{\textit{i)} Representation Collapse.} The erosion of spatial semantics destroys the topological structure of the latent space. As shown in Figure~\ref{fig:workflow}(C), critical detrimental states (e.g., agents being surrounded and losing formation) are erroneously mapped into dense clusters of successful experiences. This severe overlap between beneficial and harmful states fundamentally invalidates distance-based memory retrieval.
\textbf{\textit{ii)} Value Overestimation.} Retrieval ambiguity directly compromises incentive validity. When a pseudo-successful memory is retrieved (Figure~\ref{fig:workflow}(C), probe\textcircled{3}), existing frameworks blindly propagate its stored return as an intrinsic reward. This uncalibrated reward injection artificially inflates Q-values, creating a deceptive illusion of progress that traps agents in inescapable local optima and ultimately causes episode failure.

To resolve these coupled limitations, we propose \textbf{E}pisodic \textbf{M}emory \textbf{T}emporal \textbf{C}onsistency (\sysname{}), a unified framework that aligns robust semantic memory construction with temporally constrained incentive distribution. \sysname{} directly addresses representation collapse via a contrastive encoder-decoder architecture for precise memory retrieval, and mitigates value overestimation through a novel Bellman-consistency gating mechanism that dynamically filters misleading episodic rewards.
Our \textbf{main contributions} are summarized as follows:
\begin{itemize}[itemsep=0pt, parsep=3pt, topsep=0pt, partopsep=0pt]
  \item \textbf{Temporally Consistent Semantic Embedder:} By synergizing contrastive learning with time-conditioned reconstruction, our embedder extracts semantics-preserving state representations. This enforces a strictly clustered latent space, fundamentally resolving representation collapse and retrieval distortion in complex environments.
  \item \textbf{Temporal Consistency Gating Mechanism:} We introduce an adaptive incentive regulator grounded in the Bellman optimality criterion. By evaluating the temporal consistency error of retrieved memories, the gate dynamically attenuates rewards from pseudo-successful trajectories, effectively preventing Q-value overestimation.
  \item \textbf{Theoretical Guarantees \& Empirical Superiority:} We mathematically establish a rigorous error bound for the episodic target, formally proving that our gating mechanism safely isolates genuine optimal transitions from representation mismatch and memory staleness. Extensive evaluations on the SMAC and GRF benchmarks demonstrate that \sysname{} consistently outperforms state-of-the-art baselines, achieving absolute win-rate improvements of up to 24\% in super-hard SMAC scenarios and an average of 28\% across highly sparse GRF tasks.
\end{itemize}

\section{Background}
\label{sec:background}
\subsection{Decentralized Partially Observable Markov Decision Process}

A fully cooperative multi-agent task is formalized as a Dec-POMDP(~\cite{oliehoek2016concise}),
defined by the tuple $\mathcal{G} = \langle \mathcal{I}, \mathcal{S}, \mathcal{A}, P, R, \Omega, O, \gamma \rangle$.
Each agent $i \in \mathcal{I}$ receives a local observation $o_i = O(s, i)$ and selects action $a_i$, forming joint action $\boldsymbol{a}$ that leads to next state $s' \sim P(\cdot|s, \boldsymbol{a})$ and shared reward $r = R(s, \boldsymbol{a})$.
Due to partial observability, agents condition policies on action-observation histories $\tau_i$, with the goal of maximizing $V^{\boldsymbol{\pi}}(s) = \mathbb{E}\left[ \sum_t \gamma^t r_t \mid s_0 = s \right]$.

\subsection{Episodic Memory in MARL}
\label{subsec:episodic_marl}
Given the partial observability in Dec-POMDPs, agents must rely on historical trajectories 
for decision-making, motivating the integration of episodic memory mechanisms.
The concept is inspired by the hippocampus's capacity to rapidly encode and recall salient 
past experiences(~\cite{lee2009advances}), and in model-free RL it was formalized to circumvent 
slow parametric convergence by storing high-return trajectories(~\cite{lin2018episodic,pritzel2017neural}).

\noindent\textbf{State Representation and Memory Storage.} 
A parametric encoder $f_\phi: \mathcal{S} \to \mathbb{R}^k$ maps a global state $s_t$ to a 
compact embedding $x_t = f_\phi(s_t)$. A buffer $\mathcal{D}_E$ stores key-value pairs $(x, H(x))$, 
where $H(x)$ is the maximum empirical return. The update follows a nearest-neighbor rule:
\begin{equation}
H(x_t) = 
\begin{cases} 
\max\{H(\hat{x}_t), \mathcal{R}_t\}, & \|x_t - \hat{x}_t\|_2 < \delta \\
\mathcal{R}_t, & \text{otherwise}
\end{cases}
\end{equation}
where $\hat{x}_t = \arg\min_{x \in \mathcal{D}_E} \|x_t - x\|_2$, $\delta$ is a similarity threshold, 
and $\mathcal{R}_t$ is the cumulative return from $(s_t, a_t)$.

\noindent\textbf{Episodic Integration in MARL.} 
In cooperative MARL, episodic memory alleviates reward sparsity by recalling successful joint-action 
sequences. Early methods like EMC(~\cite{zheng2021episodic}) directly inject stored returns into the 
TD target, while EMU(~\cite{na2024efficient}) refines this by designing an explicit intrinsic reward 
$r_p$ based on the gap between the historical optimum and the current critic:
\begin{equation}
r_p \triangleq \gamma \frac{N_{\xi}(s')}{N_{\mathrm{call}}(s')} 
\left( H(f_{\phi}(s')) - \max_{a'} Q_{\theta^-}(s', a') \right)
\label{eq:rp}
\end{equation}
where $N_{\mathrm{call}}(s')$ and $N_{\xi}(s')$ count total visits and successful transitions near $s'$. 
The joint policy is optimized with a reshaped TD loss:
\begin{equation}
\mathcal{L}_{p}(\theta) =  \bigl(r(s, a) + r_p + \gamma \max_{a'} Q_{\theta^-}(s', a') - Q_{tot}(s, a; \theta)\bigr)^2
\end{equation}
where $Q_{tot}$ is the centralized joint Q-network and $Q_{\theta^-}$ is the target network.

\textbf{Limitations of Unconstrained Episodic Incentives.} Conventional episodic memory paradigms exhibit two critical coupled vulnerabilities in stochastic MARL environments. The primary bottleneck is \textit{representation collapse}, where standard deterministic reconstruction fails to enforce semantic discriminability. This distorts distance-based retrieval in $\mathcal{D}_E$, causing agents to erroneously match distinct states and recall misleading returns. Compounding this retrieval failure is the subsequent risk of \textit{value overestimation}. Derived solely from coarse, episode-level metrics, the episodic incentive $r_p$ in Eq. \ref{eq:rp} is temporally oblivious and blind to the sequential causality of individual transitions. Consequently, unregulated incentives from pseudo-successful trajectories systematically inflate Q-value estimates, trapping policies in suboptimal equilibria. Addressing this fundamental decoupling between retrieval accuracy and incentive validity necessitates a unified framework combining robust semantic representation with temporally consistent incentive gating.

\section{Method}
\label{sec:method}

\begin{figure}[t]
    \centering
    \includegraphics[width=\linewidth]{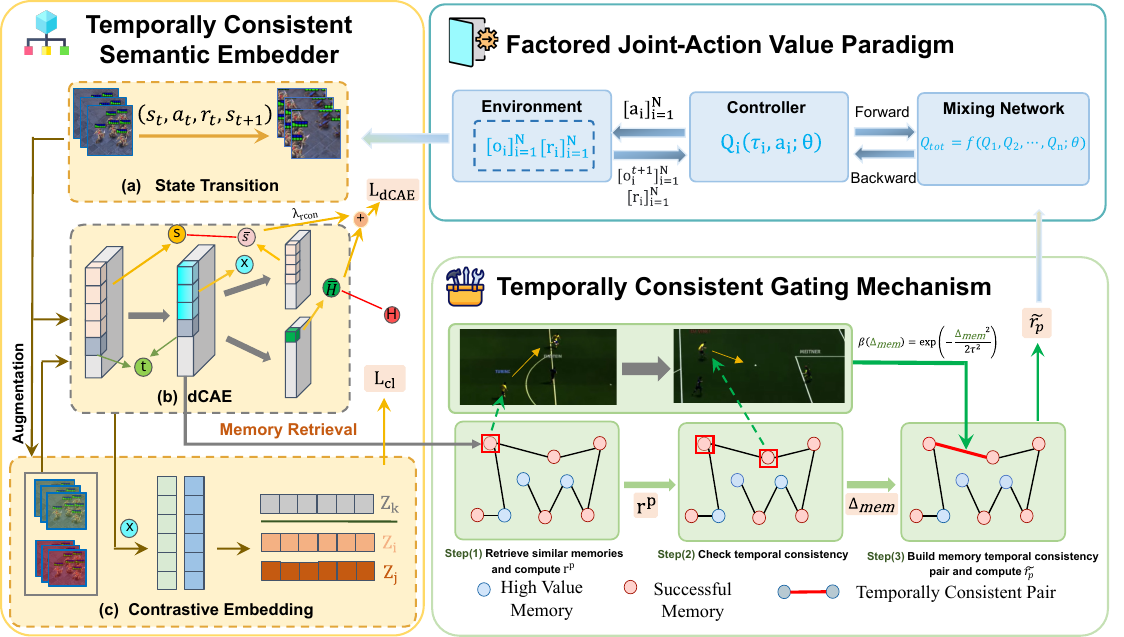}
    \caption{\textbf{Overview of the \sysname{} framework.} The architecture operates through a two-stage pipeline: (1) \textit{TCSE} regularizes the latent space via contrastive learning and time-augmented reconstruction to ensure semantically precise memory retrieval; (2) \textit{TCGM} evaluates the temporal coherence of retrieved memories using Bellman-consistency errors, dynamically scaling the episodic incentive $\tilde{r}_p$ before integration into the standard CTDE value update (denoted as the Factored Joint-Action Value Paradigm in the figure). This paradigm acts as a standard backbone (e.g., QPLEX, CDS), demonstrating that EMTC is a plug-and-play framework.}
    \label{fig:framework}
    \vspace{-15pt}
\end{figure}

This section details the architecture of \sysname{}, a unified framework designed to rectify the systemic deficiencies of unconstrained episodic guidance, as illustrated in Figure~\ref{fig:framework}. Rather than treating memory retrieval and incentive distribution as isolated modules, EMTC establishes a two-stage pipeline with periodic feedback, where latent representation quality directly governs incentive validity. The workflow operates sequentially within each training cycle: the \textit{Temporally Consistent Semantic Embedder} (TCSE) first constructs a structured latent space that preserves semantic discriminability across temporal horizons, ensuring retrieved memories are intrinsically aligned with current states. Subsequently, the \textit{Temporal Consistency Gating Mechanism} (TCGM) rigorously vets these trajectories against the Bellman optimality criterion before releasing any intrinsic reward. 
This design decouples representation learning from policy optimization while maintaining alignment through episodic buffer updates, enabling stable policy refinement even under severe reward sparsity. 

\subsection{Temporally Consistent Semantic Embedder}
\label{subsec:embedder}

\noindent\textbf{Structural Organization of Episodic Memory.} 
To ensure retrieval stability and preserve temporal dependencies, we maintain a structured episodic buffer $\mathcal{D}_E$. Each stored transition is represented as a tuple $\langle s_i, x_i, H_i, \xi_i, \text{idx}_{\text{next}} \rangle$, where $s_i$ is the raw global state, $x_i = f_\phi(s_i)$ is its latent embedding, $H_i$ records the maximum empirical return, and $\xi_i \in \{0,1\}$ denotes the trajectory's success indicator. To accommodate periodic updates to the embedding network $f_\phi$, we retain $s_i$ in the buffer, enabling on-the-fly re-projection of $x_i$ and preventing representation drift from inducing stale memory matches. Crucially, $\text{idx}_{\text{next}}$ explicitly links each entry to its immediate successor within the same trajectory. This chain-based architecture preserves the sequential causality of high-return experiences. The specific update mechanism for the episodic memory buffer is detailed in Appendix \ref{sec:appD2}.

\noindent\textbf{Multi-Objective Semantic Representation Learning.} 
We propose a dual-objective learning framework to establish a latent space where temporal consistency emerges from the synergy between semantic discriminability and temporal anchoring. As identified in Section \ref{intro} , conventional embeddings often over-enhance trajectory-level temporal trends, inadvertently blurring instantaneous spatial topologies and causing \textit{representation collapse}. To resolve this, we explicitly regularize the time-conditioned encoder $f_\phi$ and decoder $f_\psi$ via a composite loss:
\begin{equation}  \label{eq:total_loss}
\mathcal{L}_{\text{total}} = \mathcal{L}_{\text{dCAE}}(\phi, \psi) + \lambda_{\text{cl}} \mathcal{L}_{\text{cl}}(\phi),
\end{equation}
where $\lambda_{\text{cl}}$ balances the two objectives. The first component $\mathcal{L}_{\text{dCAE}}$ is a deterministic conditional autoencoder that reconstructs the state and associated highest return, conditioned on the timestep $t$:
\begin{equation} \label{eq:dcae}
\mathcal{L}_{\text{dCAE}} = \left( H_t - f_\psi^H\big(f_\phi(s_t, t), t\big) \right)^2 + \lambda_{\text{rcon}} \big\| s_t - f_\psi^s\big(f_\phi(s_t, t), t\big) \big\|_2^2,
\end{equation}
where $t$ is the conditioning timestep and $s_t$ is the global state. $H_t$ denotes the target highest return. The encoder $f_\phi$ extracts time-anchored state features, while $f_\psi^H$ and $f_\psi^s$ represent the networks for predicting the highest return and reconstructing $s_t$, respectively. $\lambda_{\text{rcon}}$ is the scale factor for the reconstruction loss. To prevent the aforementioned spatial blurring, we introduce a contrastive loss $\mathcal{L}_{\text{cl}}$ to enforce strict semantic discriminability across the temporal landscape. We adopt the InfoNCE objective~\cite{oord2018representation} to maximize mutual information between augmented views of the identical spatio-temporal state:
\begin{equation}
\mathcal{L}_{\text{cl}} = -\mathbb{E}_{(s_i,s_j)\sim\mathcal{P}} \left[ \log \frac{\exp\big(\text{sim}(z_i, z_j) / \tau_{\text{cl}}\big)}{\sum_{k=1}^N \exp\big(\text{sim}(z_i, z_k) / \tau_{\text{cl}}\big)} \right],
\end{equation}
where function $\text{sim}$ denotes cosine similarity, and $z = f_\phi(s, t)$ represents the time-conditioned latent embedding. The temperature parameter $\tau_{\text{cl}}$ controls the sharpness of the distribution. Crucially, $(z_i, z_j)$ forms a positive pair derived from two different augmented views of the same state at the same timestep (augmentation protocols detailed in Appendix~\ref{app:augmentation}).  Conversely, all other embeddings $z_k$ ($k \neq j$) originating from different timesteps or different states serve as negative samples. Algorithm \ref{alg:tcse} in Appendix~\ref{app:tcse} presents the learning framework for $f_\phi$ and $f_\psi$.

\begin{wrapfigure}{r}{0.54\textwidth}
    \vspace{-15pt}
    \centering
    \begin{subfigure}[t]{0.26\textwidth}
        \centering
        \includegraphics[width=\linewidth]{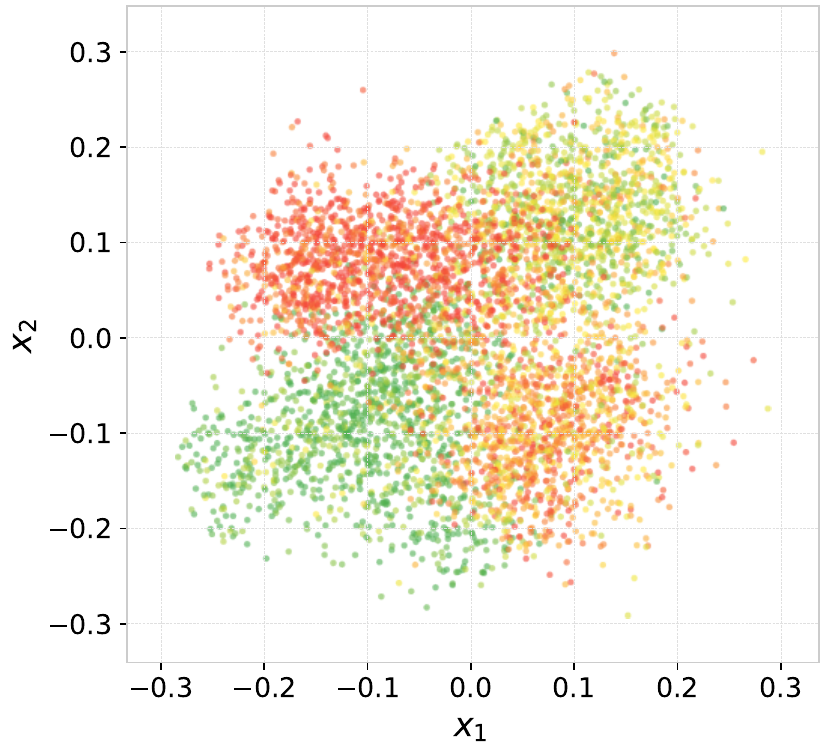}
        \caption{dCAE: overlapping clusters.}
        \label{fig:emu_collapse}
    \end{subfigure}
    \hfill
    \begin{subfigure}[t]{0.26\textwidth}
        \centering
        \includegraphics[width=\linewidth]{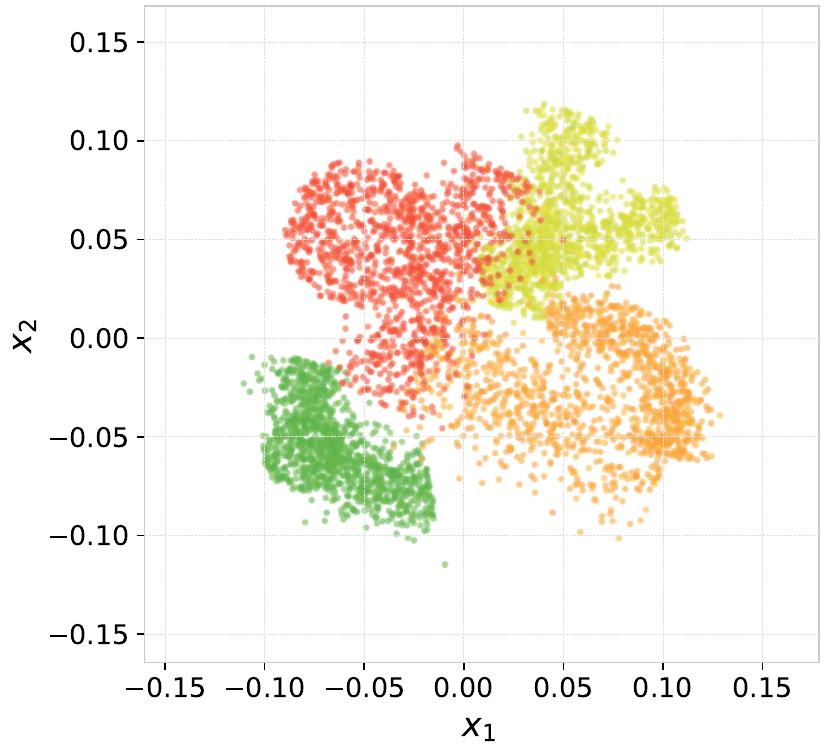}
        \caption{TCSE: compact, semantically separated clusters.}
        \label{fig:\sysname{}_clustered}
    \end{subfigure}
    \vspace{-5pt}
    \caption{Latent space comparison in the super-hard SMAC \textit{6h\_vs\_8z} scenario. The axes $x_1$ , $x_2$ denote two reduced abstract dimensions of the t-SNE embedding.}
    \label{fig:embedding_comparison}
    \vspace{-10pt}
\end{wrapfigure}

\noindent\textbf{Empirical Validation via Latent Space Visualization.} 
Figure~\ref{fig:embedding_comparison} presents a t-SNE visualization (~\cite{maaten2008visualizing}) of state embeddings $x \in \mathcal{D}_E$ sampled from the super-hard \textit{6h\_vs\_8z} scenario in SMAC. Embeddings are colored by their paired highest return $H_t$, ranging from red (low) to green (high).

As shown in Figure~\ref{fig:emu_collapse}, embeddings generated by dCAE exhibit partial semantic grouping but suffer from cluster overlap between states of differing returns, confirming the representation collapse phenomenon. In contrast, TCSE (Figure~\ref{fig:\sysname{}_clustered}) yields markedly more compact and well-separated semantic clusters. By synergizing contrastive learning with time-augmented reconstruction, our latent space effectively captures both instantaneous state features and long-term temporal progression. This enhanced clustering precision eliminates retrieval ambiguity, enabling agents to reliably leverage historical experiences for complex coordination tasks.

\subsection{Temporally Consistent Gating Mechanism}
\label{subsec:gating}

While the semantic embedder ensures precise memory retrieval, blindly distributing episodic incentives remains vulnerable to local optima when recalled trajectories lack sequential coherence. To resolve this, we introduce a \textit{Temporal Consistency Gating Mechanism} that dynamically modulates episodic rewards based on Bellman optimality. 

We first establish why temporal consistency serves as a rigorous reliability metric.

\begin{proposition}[Reliability Bound of the TCGM]
\label{prop:tcgm}
\textit{Let $V^*(s)$ be the true optimal value function. Let $\hat{V}_E(s) \doteq \mathcal{H}(\hat{s})$ denote the episodic value retrieved from the $\delta$-nearest neighbor $\hat{s} \in \mathcal{D}_E$. The empirically observed temporal consistency error $\Delta_{\text{mem}} \doteq |r + \gamma \mathcal{H}(\hat{s}') - \mathcal{H}(\hat{s})|$ is strictly bounded by:}
\begin{equation*}
    \Delta_{\text{mem}} \le \Delta_{\text{opt}} + \alpha, \quad \text{where} \quad \alpha \doteq (1+\gamma) (\epsilon_{\text{rep}} + \epsilon_{\text{store}}).
\end{equation*}
\textit{Here, $\Delta_{\text{opt}} \doteq |r + \gamma V^*(s') - V^*(s)|$ is the true underlying optimality gap of the transition. The aggregated noise term $\alpha$ consists of the representation error $\epsilon_{\text{rep}} \doteq \max_{s}|V^*(s) - V^*(\hat{s})|$ and the memory staleness $\epsilon_{\text{store}} \doteq \max_{\hat{s}}|\mathcal{H}(\hat{s}) - V^*(\hat{s})|$. }(Proof in Appendix \ref{proof:t1})

\end{proposition}

\textbf{Practical Implication.} Proposition \ref{prop:tcgm} establishes the mathematical necessity of the synergy between our two proposed components. If the semantic representation were to collapse, the noise term $\alpha$ would explode, causing the gate to falsely penalize optimal trajectories ($\Delta_{\text{opt}} \approx 0$ but $\Delta_{\text{mem}}$ is huge). However, our TCSE module explicitly minimizes $\epsilon_{\text{rep}}$ by enforcing a structured latent space. By tightly bounding the noise $\alpha$, EMTC guarantees that a large $\Delta_{\text{mem}}$ strictly implies a large $\Delta_{\text{opt}}$. Thus, the gating mechanism acts as a safe and unbiased filter, effectively decoupling true exploration signals from representation artifacts.

\noindent\textbf{Mechanism Workflow \& Formulation.} Building on this insight, the gating mechanism operates through a closed-loop pipeline that directly leverages the structured buffer from Section \ref{subsec:embedder}:
\begin{enumerate}[leftmargin=*, label=(\roman*)] 
    \item \textbf{Consistency Error Computation:} For each transition $(s, a)$, we retrieve the successor's stored value $\mathcal{H}(s')$ via the structural link $\text{idx}_{\text{next}}$ in $\mathcal{D}_E$. We compute the \textit{Temporal Consistency Error}:
    \begin{equation}
        \Delta_{\text{mem}}(s, a) = \left| r(s,a) + \gamma \mathcal{H}(s') - \mathcal{H}(s) \right|,
        \label{eq:delta_mem}
    \end{equation}
    which quantifies the discrepancy between the one-step Bellman update and the stored memory value.
    
    \item \textbf{Adaptive Gating:} To translate this error into an incentive constraint, we define the \textit{Consistency Coefficient} $\beta(\Delta_{\text{mem}})$ as a Gaussian modulation function:
    \begin{equation}
        \beta(\Delta_{\text{mem}}) = \exp \left( -\frac{\Delta_{\text{mem}}(s, a)^2}{2\tau^2} \right),
        \label{eq:beta_gate}
    \end{equation}
    where $\tau$ is a temperature parameter (distinct from the contrastive learning temperature $\tau_{cl}$). Our exhaustive hyperparameter experiments reveal that different values of $\tau$ exert distinct impacts on the learning dynamics across the early and late stages of training. Motivated by these findings, we adopt a linear annealing strategy for $\tau$ (see Appendix~\ref{app:tau} for details). Since $\beta \in (0, 1]$, highly inconsistent transitions ($\Delta_{\text{mem}} \gg 0$) are heavily attenuated ($\beta \to 0$), automatically suppressing misleading incentives.
    
    \item \textbf{Gated Reward \& Loss Integration:} The coefficient dynamically scales the baseline episodic incentive $r_p$ (Eq. ~\eqref{eq:rp}) to produce the \textit{Consistency-Aware Episodic Reward} $\tilde{r}_p$:
    \begin{equation}
        \tilde{r}_p(s, a) = \beta(\Delta_{\text{mem}}) \cdot r_p(s, a) = \beta(\Delta_{\text{mem}}) \cdot \gamma \frac{N_\xi(s')}{N_{\text{call}}(s')} \left( \mathcal{H}(s') - \max_{a'} Q_{\theta^-}(s', a') \right).
        \label{eq:gated_rp}
    \end{equation}
    Integrating $\tilde{r}_p$ into the value factorization objective yields the gated TD loss:
    \begin{equation}
        \mathcal{L}_\theta^\beta =( r + \tilde{r}_p + \gamma \max_{a'} Q_{\theta^-}(s', a') - Q_{tot}(s, a; \theta) )^2 
        \label{eq:gated_loss}
    \end{equation}
\end{enumerate}

\noindent\textbf{Gradient Analysis \& Regulatory Behavior.} The gradient of $\mathcal{L}_\theta^\beta$ with respect to $\theta$ reveals the mechanism's regulatory dynamics:
\begin{equation}
\begin{aligned}
    \nabla_\theta \mathcal{L}_\theta^\beta &= -2 \nabla_\theta Q_{tot}(s, a; \theta) \left( \Delta_{\text{TD}} + \tilde{r}_p \right) \\
    &= -2 \nabla_\theta Q_{tot}(s, a; \theta) \left( \Delta_{\text{TD}} + \beta(\Delta_{\text{mem}}) \cdot \gamma \frac{N_\xi(s')}{N_{\text{call}}(s')} \eta_{\text{val}}(s') \right),
\end{aligned}
\label{eq:gradient}
\end{equation}
where $\Delta_{\text{TD}} = r + \gamma \max_{a'} Q_{\theta^-}(s', a') - Q_{tot}(s, a; \theta)$ is the standard TD error, and $\eta_{\text{val}}(s') \doteq \mathcal{H}(s') - \max_{a'} Q_{\theta^-}(s', a')$ is the episodic value gap. Here, the gradient signal $\nabla_\theta \mathcal{L}^\beta_\theta$ with the proposed \textit{Consistency-Aware Episodic Reward} $\tilde{r}_p$ can accurately estimate the optimal gradient signal as follows.

\begin{proposition}[Asymptotic Unbiasedness]
\label{prop:asymp}
\textit{Let $\nabla_\theta \mathcal{L}^\beta_\theta$ be the gradient induced by the gated reward. 
Assuming the policy $\pi_\theta$ approaches optimality and the episodic memory matures 
($\pi_\theta \to \pi^*$, $\mathcal{H} \to V^*$), the gated gradient accurately recovers the 
true optimal gradient: 
$\lim_{\pi_\theta \to \pi^*,\; \mathcal{H} \to V^*} \nabla_\theta \mathcal{L}^\beta_\theta = \nabla_\theta \mathcal{L}^*_\theta$.} 
(Proof in Appendix~\ref{proof:t2})
\end{proposition}

Proposition \ref{prop:asymp}  confirms that our mechanism functions as a \textit{safe regularizer} with a dual-phase property that directly addresses the limitations of unconstrained episodic MARL:
\begin{itemize}[leftmargin=*, nosep]
    \item \textbf{Early-Stage Lenient Exploration:} During initial training, $\Delta_{\text{mem}}$ is inherently large due to disjointed trajectories. To prevent exploration deadlocks, we utilize a higher initial temperature $\tau$. This lenient gating allows moderately consistent episodic incentives to pass, actively encouraging rapid and directed early exploration.
    \item \textbf{Late-Stage Strict Filtering \& Convergence:} As training advances, $\tau$ linearly decays to impose a stricter consistency bound, rigorously filtering out spurious pseudo-memories to prevent Q-value overestimation. Concurrently, as high-quality experiences accumulate and $\mathcal{H} \to V^*$, genuine optimal memories yield $\Delta_{\text{mem}} \to 0$ and $\beta \to 1$. The mechanism naturally anneals to an unbiased guidance signal, ensuring the optimization trajectory is never perturbed away from the true optimum.
   
\end{itemize}
This dual-phase regulation enables \sysname{} to maintain stable, directed exploration in super-hard scenarios where traditional episodic incentives typically collapse. The complete gating procedure and its integration into the training loop are detailed in Appendix~\ref{app:main}.

\section{Experiments}
\label{sec:experiments}

We design our experimental evaluation to systematically address three core research questions:
\begin{itemize}[nosep, leftmargin=*, topsep=3pt]
    \item \textbf{Q1 (Overall Efficacy):} How does \sysname{} compare to state-of-the-art MARL frameworks?
    \item \textbf{Q2 (Embedder Contribution):} To what extent does the Temporally Consistent Semantic Embedder (TCSE) improve retrieval precision and latent space discriminability?
    \item \textbf{Q3 (Gating Mechanism):} How does the Temporal Consistency Gating Mechanism (TCGM) regulate episodic incentives to ensure stable policy refinement?
\end{itemize}

\noindent\textbf{Benchmarks.} 
We evaluate \sysname{} on two widely-adopted MARL benchmarks that jointly cover diverse coordination challenges: 
(1) \textbf{StarCraft Multi-Agent Challenge (SMAC)}(~\cite{samvelyan2019starcraft}), featuring heterogeneous unit types, partial observability, and combinatorial action spaces; 
(2) \textbf{Google Research Football (GRF)}(~\cite{kurach2020google}), characterized by continuous-state dynamics, sparse goal-based rewards, and long-horizon strategic planning. 
This dual-domain setup enables comprehensive assessment of generalization across discrete/continuous state spaces and dense/sparse reward regimes. The details of the benchmarks are presented in Appendix \ref{app:environment}.

\noindent\textbf{Baselines and Implementations.}
Our primary comparison target is \textbf{EMU}(~\cite{na2024efficient}), the current state-of-the-art episodic memory framework in MARL. 
To contextualize performance gains, we additionally include strong value-based baselines: \textbf{QMIX}(~\cite{rashid2020monotonic}), \textbf{QPLEX}(~\cite{wang2020qplex}), and \textbf{CDS}(~\cite{li2021celebrating}). 
Since \sysname{} is architecture-agnostic and plugs into any CTDE value-factorization backbone, we report results for two instantiations: \textbf{\sysname{}(QPLEX)} and \textbf{\sysname{}(CDS)}. Appendix \ref{app:experiment details} and \ref{appendix:infrastructure} provides further details of experiment settings
and computational overhead. And additional performance comparisons demonstrating \sysname{}'s superiority over MAPPO (\cite{yu2022surprising}) and DMIX (\cite{sun2021dfac}) are provided in Appendix~\ref{app:vs_mappo}.

\begin{figure}[H]
    \centering
    \includegraphics[width=0.8\linewidth]{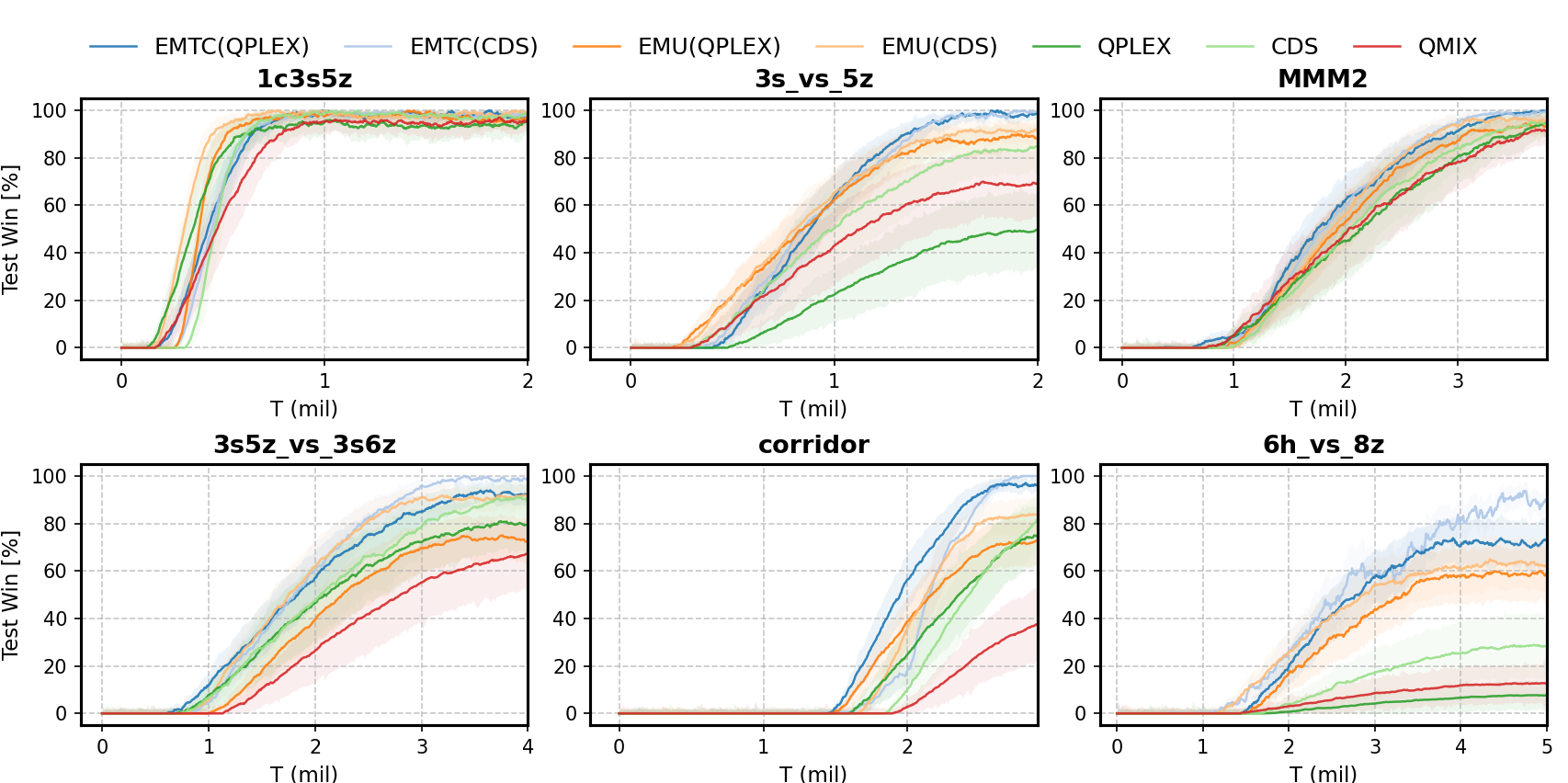}
    \caption{Performance comparison of \sysname{} against EMU and other state-of-the-art baselines on SMAC scenarios. The evaluation includes easy and hard maps (1c3s5z, 3s\_vs\_5z) and super-hard maps (MMM2, 3s5z\_vs\_3s6z, corridor, and 6h\_vs\_8z). Notably, T represents the training time in millions of steps in all figures. }
    \label{fig:smac}
    \vspace{-1.5\baselineskip}
\end{figure}

\subsection{Q1. Evaluation on StarCraft II (SMAC)}
\label{sec:exp_smac}

Figure~\ref{fig:smac} presents win-rate curves across six SMAC scenarios of increasing difficulty. Both \sysname{}(QPLEX) and \sysname{}(CDS) consistently surpass their backbone architectures and the episodic baseline EMU. Especially in super-hard scenarios (corridor, 6h\_vs\_8z, 3s5z\_vs\_3s6z), while maintaining almost the same convergence speed, \sysname{} achieves an average win-rate gain of 24\%.

\subsection{Q1. Evaluation on Google Research Football (GRF)}
\label{sec:exp_grf}

\begin{figure}[H]
    \centering
    \includegraphics[width=0.8\linewidth]{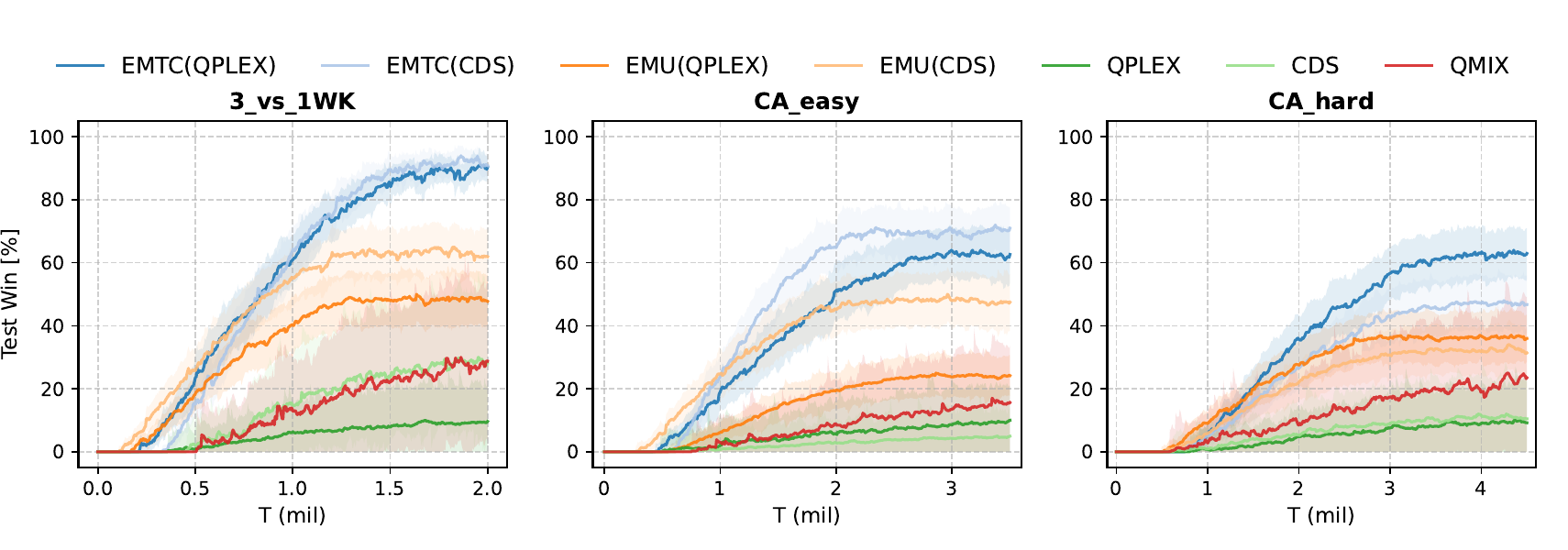}
    \caption{Performance comparison of \sysname{} against baseline algorithms on GRF.}
    \label{fig:grf}
    \vspace{-1\baselineskip}
\end{figure}

Here, we evaluate \sysname{} on GRF, which features continuous-state dynamics and extremely sparse goal-based rewards, i.e., a setting markedly different from SMAC. 
As shown in Figure~\ref{fig:grf}, \sysname{} maintains its advantage over EMU and value-based baselines, delivering an average win-rate improvement of 28\%. 
Notably, the same two mechanisms (TCSE + TCGM) that proved effective in SMAC here preserve spatial semantics under partial observability and filter inconsistent incentives in sparse-reward regimes. 
This cross-environment consistency confirms that temporally constrained episodic guidance generalizes beyond discrete-action benchmarks to complex, real-world-inspired coordination tasks. See Appendix \ref{app:quantitative_results} for detailed quantitative results.

\subsection{Q2. Ablation Study on TCSE}
\label{sec:exp_embedder}

To isolate the contribution of the semantic embedder to retrieval precision (Q2), we evaluate three designs, including random projection (\cite{dasgupta2003elementary}), dCAE (Eq.~\ref{eq:dcae}), and TCSE (Eq.~\ref{eq:total_loss}), using both final win-rate and the composite efficiency metric $\bar{\mu}_w$ (Appendix \ref{app:uw}). All evaluations disable the TCGM ($\beta$) to decouple embedding quality from incentive regulation, and vary the retrieval radius $\delta$ across logarithmic scales.

\begin{figure*}[htbp]
    \centering
    \begin{minipage}[t]{0.485\textwidth}
        \centering
        \begin{subfigure}[b]{0.48\linewidth}
            \includegraphics[width=\linewidth]{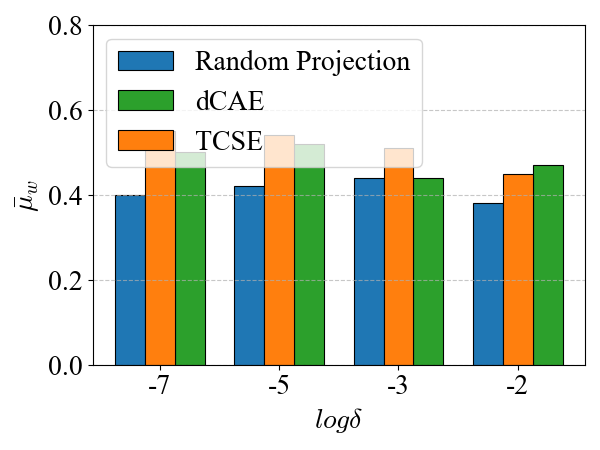}
            \caption{3s\_vs\_5z}
        \end{subfigure}
        \hfill
        \begin{subfigure}[b]{0.48\linewidth}
            \includegraphics[width=\linewidth]{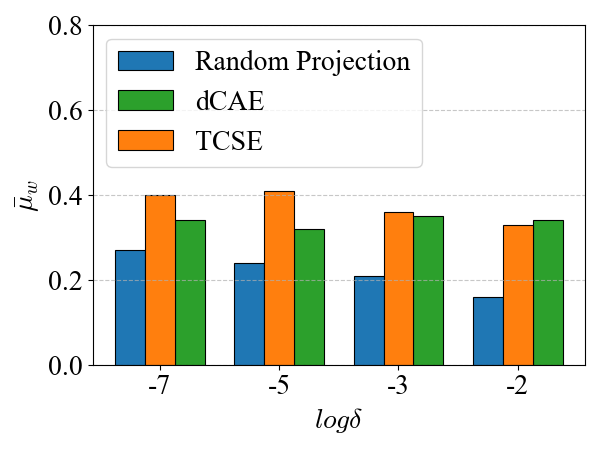}
            \caption{6h\_vs\_8z}
        \end{subfigure}
        \vspace{0.2cm}
        \caption{$\bar{\mu}_w$ values according to different $\delta$.}
        \label{fig:mu_w}
    \end{minipage}
    \hfill 
    \begin{minipage}[t]{0.485\textwidth}
        \centering
        \begin{subfigure}[b]{0.48\linewidth}
            \includegraphics[width=\linewidth]{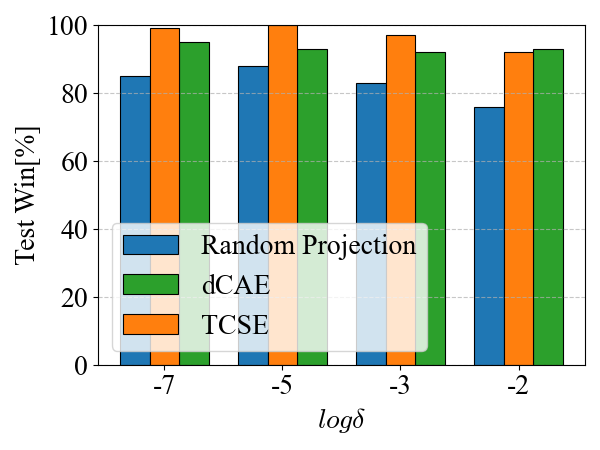}
            \caption{3s\_vs\_5z}
        \end{subfigure}
        \hfill
        \begin{subfigure}[b]{0.48\linewidth}
            \includegraphics[width=\linewidth]{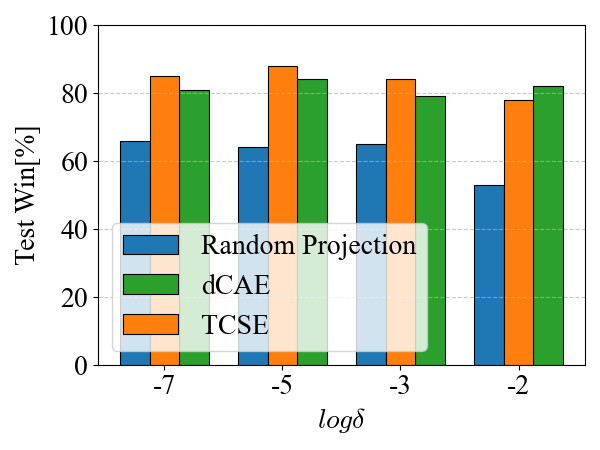}
            \caption{6h\_vs\_8z}
        \end{subfigure}
        \vspace{0.2cm}
        \caption{Test win-rate according to different $\delta$.}
        \label{fig:test_win_rate}
    \end{minipage}
\end{figure*}

Figures~\ref{fig:mu_w} and~\ref{fig:test_win_rate} report $\bar{\mu}_w$ and final test win-rates under varying $\delta$ on SMAC scenarios. More data of $\bar{\mu}_w$ is presented in Tables \ref{tab:embed_3s5z} and \ref{tab:embed_corridor} in Appendix \ref{app:ablation_emb}. As shown, TCSE consistently achieves higher $\bar{\mu}_w$ and win-rates across all $\delta$ settings on both easy (3s\_vs\_5z) and super-hard (6h\_vs\_8z) maps. Crucially, TCSE exhibits markedly lower sensitivity to $\delta$ fluctuations compared to dCAE and random projection. We attribute this robustness to TCSE's dual-objective design, i.e., contrastive learning enforces semantic clustering to reduce retrieval ambiguity, while time-conditioned reconstruction preserves temporal anchors for coherent trajectory matching. More parametric study is in Appendix \ref{app:Parametric Study on TCSE}

\subsection{Q2 \& Q3. Further Ablation Study}

\begin{figure*}[htbp] 
    \centering
    \begin{subfigure}{0.32\textwidth}
        \centering
        \includegraphics[width=\linewidth]{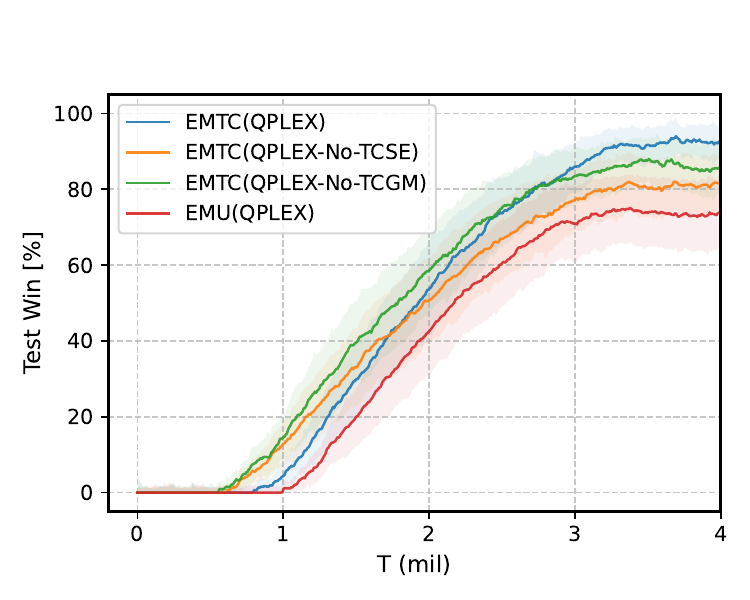}
        \caption{\textit{3s5z\_vs\_3s6z}}
        \label{fig:ablation_3s5z}
    \end{subfigure}
    \hfill
    \begin{subfigure}{0.32\textwidth}
        \centering
        \includegraphics[width=\linewidth]{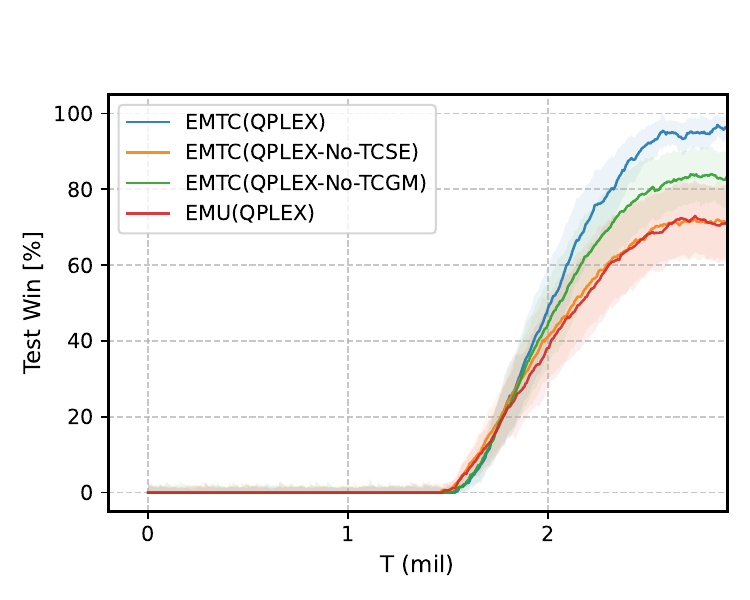}
        \caption{\textit{corridor}}
        \label{fig:ablation_corridor}
    \end{subfigure}
    \hfill
    \begin{subfigure}{0.32\textwidth}
        \centering
        \includegraphics[width=\linewidth]{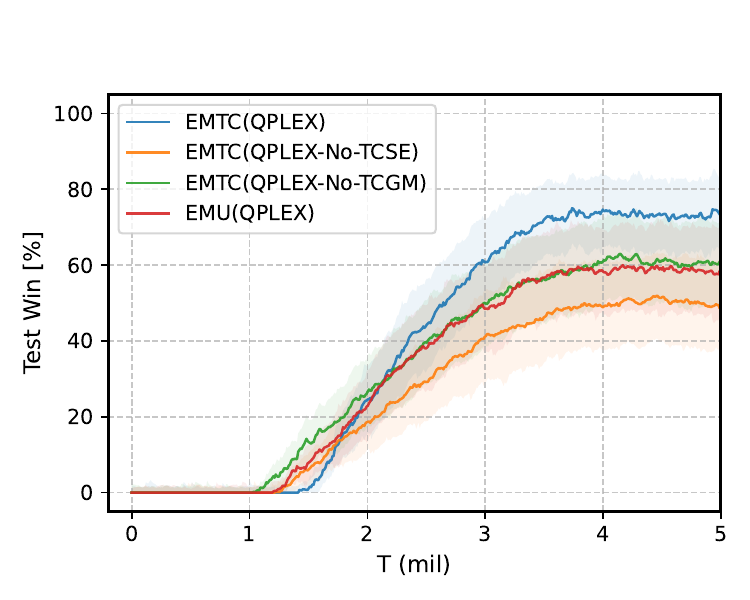}
        \caption{\textit{6h\_vs\_8z}}
        \label{fig:ablation_6h}
    \end{subfigure}

    \vspace{-3pt} 
    \caption{Further ablation studies on complex MARL tasks.}
    \label{fig:ablation_study}
\end{figure*}

We conduct comprehensive ablation studies to evaluate the individual contributions of the core components. Starting from the full \textbf{\sysname{} (QPLEX)} model, we remove the gating mechanism to investigate the effect of unregulated incentives, denoted as \textbf{\sysname{} (QPLEX-No-TCGM)}. We additionally ablate the TCSE component, denoted as \textbf{\sysname{} (QPLEX-No-TCSE)}, to assess the impact of representation quality on memory retrieval efficiency. Furthermore, we include \textbf{EMU (QPLEX)} as a prominent baseline to highlight the performance gains achieved by our temporal consistency constraints. The performance of each variant is evaluated on three super-hard SMAC scenarios: \textit{3s5z\_vs\_3s6z}, \textit{corridor}, and \textit{6h\_vs\_8z}. As illustrated in Figure \ref{fig:ablation_study}, the experimental results demonstrate that the removal of either component leads to a significant degradation in win rates, underscoring the synergy between robust embedding and rigorous incentive regulation.

\subsection{Q3. Qualitative Analysis on TCGM}
\vspace{-1cm}

\begin{wrapfigure}{r}{0.6\textwidth}
    \vspace{-10pt}
    \centering
    \begin{subfigure}{\linewidth}
        \centering
        \includegraphics[width=\linewidth]{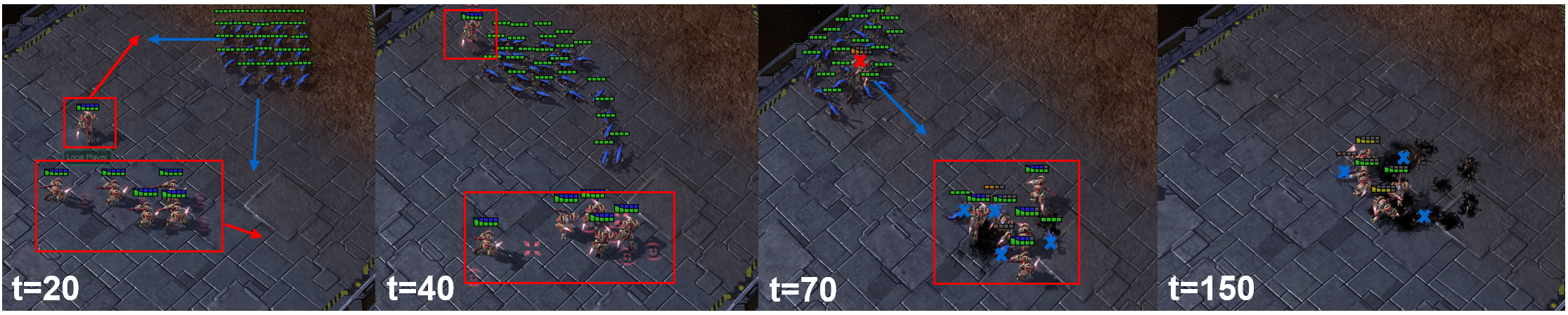}
        \caption{Effective 1-5 asymmetrical split (Bait-and-encircle)}
        \label{fig:strategy_good}
    \end{subfigure}
    \\[5pt]
    \begin{subfigure}{\linewidth}
        \centering
        \includegraphics[width=\linewidth]{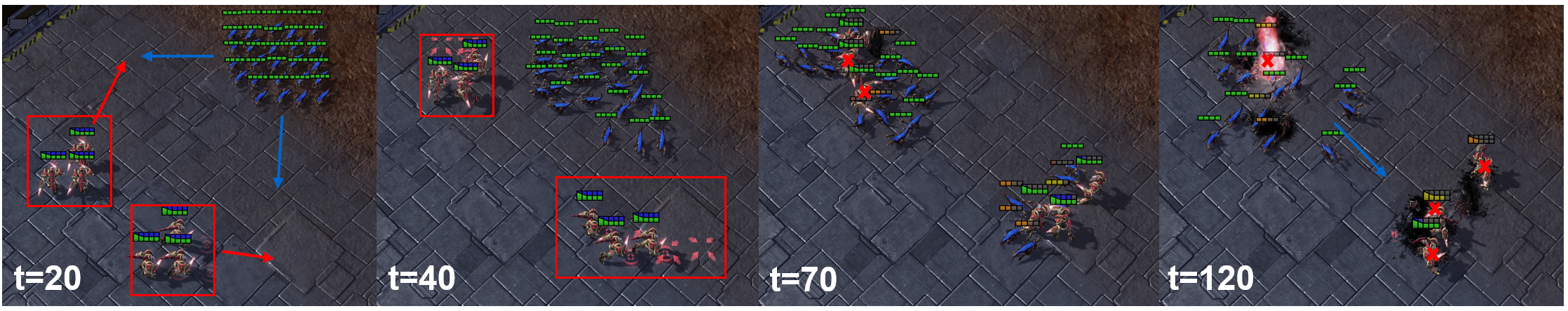}
        \caption{Sub-optimal 3-3 even split}
        \label{fig:strategy_bad}
    \end{subfigure}
    \\[10pt]
    \begin{subfigure}{0.48\linewidth}
        \centering
        \includegraphics[width=\linewidth]{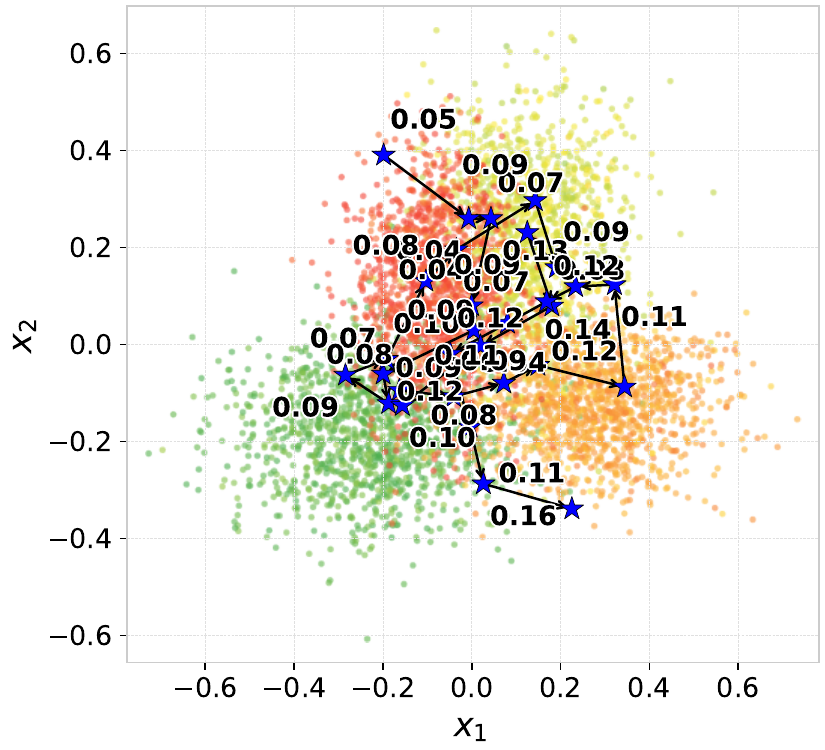}
        \caption{Good Strategy (1-5 split)}
        \label{fig:tsne_good}
    \end{subfigure}
    \hfill
    \begin{subfigure}{0.48\linewidth}
        \centering
        \includegraphics[width=\linewidth]{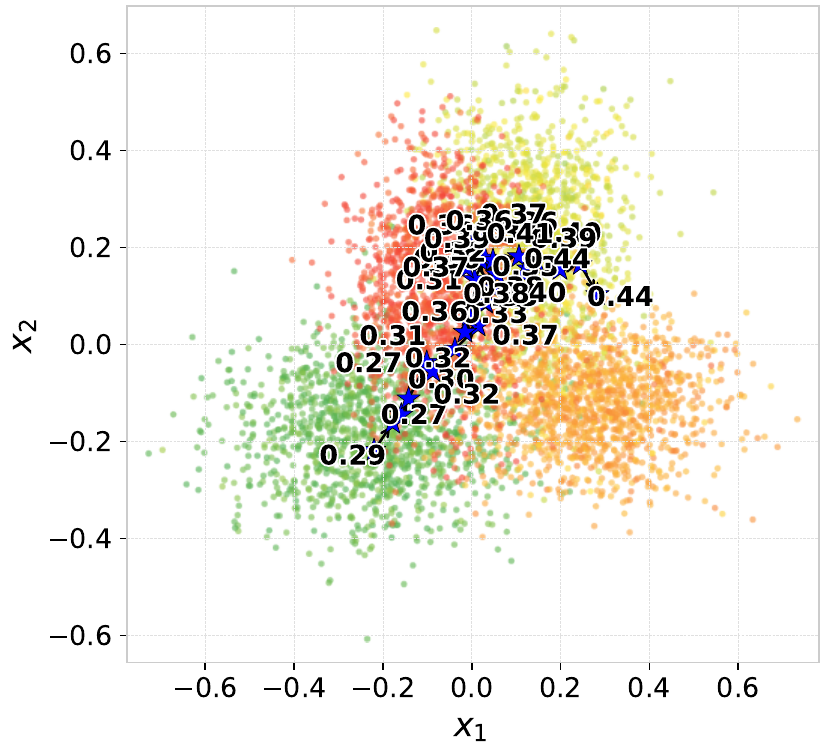}
        \caption{Bad Strategy (3-3 split)}
        \label{fig:tsne_bad}
    \end{subfigure}
    \caption{Overall analysis in \textit{corridor}. (a)\&(b) Strategy divergence; (c)\&(d) t-SNE projections of the corresponding memory retrieval trajectories. The axes $x_1$ and $x_2$ denote the two reduced abstract dimensions of the t-SNE embedding.}
    \label{fig:replay_vis}
    \vspace{-10pt}
\end{wrapfigure}

We qualitatively investigate how the Temporal Consistency Gating Mechanism (TCGM) regulates policy learning by visualizing the gameplay dynamics alongside the latent memory buffer $\mathcal{D}_E$. We conduct our analysis on the challenging \textit{corridor} scenario, where agents must navigate complex spatial constraints. During exploration, two strategic patterns emerge: a successful asymmetrical 1-5 split, where one agent baits enemies while the others encircle and eventually eliminate all opponents (Figure~\ref{fig:strategy_good}); and a sub-optimal even 3-3 split, which lacks sufficient concentrated firepower and consistently fails to defeat the enemies (Figure~\ref{fig:strategy_bad}).

To understand TCGM's evaluation, we project $\mathcal{D}_E$ via t-SNE and trace 
memory retrievals during the first 200 steps (sampled every 10 steps), where the strategic 
divergence occurs. As shown in Figures~\ref{fig:tsne_good} and \ref{fig:tsne_bad}, the 
superior 1-5 split traverses a smooth latent region with consistently low $\Delta_{\text{mem}}$, 
prompting TCGM to assign $\beta \to 1$ and preserve the episodic incentive. Conversely, 
the flawed 3-3 split retrieves disjointed memories with dramatically larger $\Delta_{\text{mem}}$ 
(annotated in Figure~\ref{fig:tsne_bad}), causing TCGM to aggressively attenuate the reward 
($\beta \to 0$) and shield the policy from misleading guidance. This contrast validates 
that TCGM acts as a temporally-grounded safeguard, propagating episodic incentives only when 
they are reliable. Appendix~\ref{app:delta_beta_dynamics} details the training dynamics of 
$\Delta_{\text{mem}}$ and $\beta$.

\section{Conclusion}
In this paper, we address the severe reward sparsity and limitations of traditional episodic memory in cooperative Multi-Agent Reinforcement Learning (MARL). To prevent agents from being trapped in local optima caused by blind incentive distributions, we propose Episodic Memory Temporal Consistency (EMTC). EMTC utilizes a contrastive learning-enhanced architecture to construct a temporally consistent semantic embedder, and introduces a novel gating mechanism to selectively distribute incentives. This approach effectively filters out misleading signals and mitigates TD target overestimation in Q-learning. Supported by theoretical guarantees, extensive evaluations on the GRF and SMAC benchmarks demonstrate that EMTC significantly outperforms conventional paradigms. A detailed discussion of the limitations and future directions of this work is provided in Appendix~\ref{sec:limitations}.

\clearpage
\newpage
\bibliographystyle{abbrvnat}
\bibliography{references}

@article{zhou2022multi,
  title={Multi-agent reinforcement learning for cooperative lane changing of connected and autonomous vehicles in mixed traffic},
  author={Zhou, W. and Chen, D. and Yan, J. and others},
  journal={Autonomous Intelligent Systems},
  volume={2},
  number={1},
  pages={5},
  year={2022}
}

@inproceedings{hong2024multi,
  title={Multi-agent collaborative perception via motion-aware robust communication network},
  author={Hong, S. and Liu, Y. and Li, Z. and others},
  booktitle={Proceedings of the IEEE/CVF Conference on Computer Vision and Pattern Recognition},
  pages={15301--15310},
  year={2024}
}

@article{oliehoek2008optimal,
  title={Optimal and approximate {Q}-value functions for decentralized {POMDP}s},
  author={Oliehoek, Frans A and Spaan, Matthijs TJ and Vlassis, Nikos},
  journal={Journal of Artificial Intelligence Research},
  volume={32},
  pages={289--353},
  year={2008}
}

@book{oliehoek2016concise,
  title={A concise introduction to decentralized {POMDP}s},
  author={Oliehoek, Frans A and Amato, Christopher},
  publisher={Springer International Publishing},
  address={Cham, Switzerland},
  year={2016}
}

@inproceedings{gupta2017cooperative,
  title={Cooperative multi-agent control using deep reinforcement learning},
  author={Gupta, Jayesh K and Egorov, Maxim and Kochenderfer, Mykel},
  booktitle={International conference on autonomous agents and multiagent systems},
  pages={66--83},
  year={2017},
  organization={Springer}
}

@article{rashid2020monotonic,
  title={Monotonic value function factorisation for deep multi-agent reinforcement learning},
  author={Rashid, Tabish and Samvelyan, Mikayel and De Witt, Christian Schroeder and others},
  journal={Journal of Machine Learning Research},
  volume={21},
  number={178},
  pages={1--51},
  year={2020}
}

@inproceedings{sun2021dfac,
  title={{DFAC} framework: Factorizing the value function via quantile mixture for multi-agent distributional {Q}-learning},
  author={Sun, Wei-Fang and Lee, Cheng-Kuang and Lee, Chun-Yi},
  booktitle={International conference on machine learning},
  pages={9945--9954},
  year={2021},
  organization={PMLR}
}

@article{wang2020qplex,
  title={{Q}plex: Duplex dueling multi-agent {Q}-learning},
  author={Wang, Jianhao and Ren, Zhizhou and Liu, Terry and others},
  journal={arXiv preprint arXiv:2008.01062},
  year={2020}
}

@article{blundell2016model,
  title={Model-free episodic control},
  author={Blundell, Charles and others},
  journal={arXiv preprint arXiv:1606.04460},
  year={2016}
}

@article{zheng2021episodic,
  title={Episodic multi-agent reinforcement learning with curiosity-driven exploration},
  author={Zheng, L. and Chen, J. and Wang, J. and others},
  journal={Advances in Neural Information Processing Systems},
  volume={34},
  pages={3757--3769},
  year={2021}
}

@article{na2024efficient,
  title={Efficient episodic memory utilization of cooperative multi-agent reinforcement learning},
  author={Na, H. and Seo, Y. and Moon, I.},
  journal={arXiv preprint arXiv:2403.01112},
  year={2024}
}

@article{lengyel2007hippocampal,
  title={Hippocampal contributions to control: the third way},
  author={Lengyel, M{\'a}t{\'e} and Dayan, Peter},
  journal={Advances in neural information processing systems},
  volume={20},
  year={2007}
}

@article{lin2018episodic,
  title={Episodic memory deep {Q}-networks},
  author={Lin, Zichuan and Zhao, T. and Yang, G. and others},
  journal={arXiv preprint arXiv:1805.07603},
  year={2018}
}

@inproceedings{pritzel2017neural,
  title={Neural episodic control},
  author={Pritzel, Alexander and Uria, B. and Srinivasan, S. and others},
  booktitle={International conference on machine learning},
  pages={2827--2836},
  year={2017},
  organization={PMLR}
}

@book{sutton1998reinforcement,
  title={Reinforcement learning: An introduction},
  author={Sutton, R. S. and Barto, A. G.},
  publisher={MIT press},
  address={Cambridge},
  year={1998}
}

@incollection{barto2012intrinsic,
  title={Intrinsic motivation and reinforcement learning},
  author={Barto, A. G.},
  booktitle={Intrinsically motivated learning in natural and artificial systems},
  pages={17--47},
  year={2012},
  publisher={Springer}
}

@inproceedings{van2016deep,
  title={Deep Reinforcement Learning with Double Q-Learning},
  author={Van Hasselt, Hado and Guez, Arthur and Silver, David},
  booktitle={Proceedings of the Thirtieth AAAI Conference on Artificial Intelligence (AAAI)},
  pages={2094--2100},
  year={2016}
}

@inproceedings{ng1999policy,
  title={Policy invariance under reward transformations: Theory and application to reward shaping},
  author={Ng, A. Y. and Harada, D. and Russell, S.},
  booktitle={ICML},
  volume={99},
  pages={278--287},
  year={1999}
}

@inproceedings{fujimoto2018addressing,
  title={Addressing function approximation error in actor-critic methods},
  author={Fujimoto, S. and Hoof, H. and Meger, D.},
  booktitle={International conference on machine learning},
  pages={1587--1596},
  year={2018},
  organization={PMLR}
}

@inproceedings{henderson2018deep,
  title={Deep Reinforcement Learning That Matters},
  author={Henderson, Peter and Islam, Riashat and Bachman, Philip and Pineau, Joelle and Precup, Doina and Meger, David},
  booktitle={Proceedings of the Thirty-Second AAAI Conference on Artificial Intelligence (AAAI)},
  pages={3207--3214},
  year={2018}
}

@article{bellemare2016unifying,
  title={Unifying count-based exploration and intrinsic motivation},
  author={Bellemare, M. and Srinivasan, S. and Ostrovski, G. and others},
  journal={Advances in neural information processing systems},
  volume={29},
  year={2016}
}

@techreport{lee2009advances,
  title={Advances in neural information processing systems 22},
  author={Lee, D. D. and Pham, P. and Largman, Y. and others},
  year={2009},
  institution={Tech Rep}
}

@article{oord2018representation,
  title={Representation learning with contrastive predictive coding},
  author={van den Oord, A{\"a}ron and Li, Yazhe and Vinyals, Oriol},
  journal={arXiv preprint arXiv:1807.03748},
  year={2018}
}

@article{maaten2008visualizing,
  title={Visualizing data using t-SNE},
  author={Van der Maaten, Laurens and Hinton, Geoffrey},
  journal={Journal of Machine Learning Research},
  volume={9},
  number={11},
  year={2008}
}

@article{samvelyan2019starcraft,
  title={The {S}tar{C}raft multi-agent challenge},
  author={Samvelyan, Mikayel and Rashid, Tabish and De Witt, Christian Schroeder and Farquhar, Gregory and Nardelli, Nantas and Peng, Bei and Cassici, Lavine and Foerster, Jakob and Whiteson, Shimon},
  journal={arXiv preprint arXiv:1902.04043},
  year={2019}
}

@inproceedings{kurach2020google,
  title={Google research football: A novel reinforcement learning environment},
  author={Kurach, Karol and Raichuk, Anton and Sta{\'n}czyk, Piotr and Zaj{\k{a}}c, Micha{\l} and Gelly, Sylvain and David, Lasse and Zapala, Mateusz and Li, Adam and Sedghi, Hanie and Kleineberg, Katja and others},
  booktitle={Proceedings of the AAAI Conference on Artificial Intelligence},
  year={2020}
}

@inproceedings{li2021celebrating,
  title={Celebrating Diversity in Shared Multi-Agent Reinforcement Learning},
  author={Li, Chenghao and Wang, Tonghan and Zheng, Chengjie and Zhu, Jiaming and Yao, Junyang and Zhao, Xiangyang and Zhang, Chongjie},
  booktitle={Advances in Neural Information Processing Systems (NeurIPS)},
  year={2021}
}

@article{yu2022surprising,
  title={The surprising effectiveness of ppo in cooperative multi-agent games},
  author={Yu, Chao and Velu, Akash and Vinitsky, Eugene and Gao, Jiaxuan and Wang, Yu and Bayen, Alexandre and Wu, Yi},
  journal={Advances in neural information processing systems},
  volume={35},
  pages={24611--24624},
  year={2022}
}

@article{dasgupta2003elementary,
  title={An elementary proof of a theorem of Johnson and Lindenstrauss},
  author={Dasgupta, Sanjoy and Gupta, Anupam},
  journal={Random Structures \& Algorithms},
  volume={22},
  number={1},
  pages={60--65},
  year={2003},
  publisher={Wiley Online Library},
  doi={10.1002/rsa.10073}
}

@inproceedings{wang2021rode,
  title     = {{RODE}: Learning Roles to Decompose Multi-Agent Tasks},
  author    = {Wang, Tonghan and Gupta, Tarun and Mahajan, Anuj and Peng, Bei and Whiteson, Shimon and Zhang, Chongjie},
  booktitle = {Proceedings of the International Conference on Learning Representations (ICLR)},
  year      = {2021}
}

@inproceedings{luo2024euclidean,
  title={Reinforcement Learning with Euclidean Data Augmentation for State-Based Continuous Control},
  author={Luo, Jinzhu and Chen, Dingyang and Zhang, Qi},
  booktitle={Advances in Neural Information Processing Systems (NeurIPS)},
  year={2024}
}

@inproceedings{adaptaug2024,
  title={AdaptAUG: Adaptive Data Augmentation Framework for Multi-Agent Reinforcement Learning},
  author={Zhang, Y. } ,
  booktitle={2024 IEEE International Conference on Robotics and Automation (ICRA)},
  year={2024}
}

@inproceedings{yu2023esp,
  title={ESP: Exploiting Symmetry Prior for Multi-Agent Reinforcement Learning},
  author={Yu, Xin and Shi, Rongye and Feng, Pu and Tian, Yongkai and Luo, Jie and Wu, Wenjun},
  booktitle={26th European Conference on Artificial Intelligence (ECAI)},
  year={2023}
}

@inproceedings{chen2025multi,
  author    = {Junhong Chen and others},
  title     = {Multi-agent systems for robotic autonomy with {LLMs}},
  booktitle = {Proceedings of the IEEE/CVF Conference on Computer Vision and Pattern Recognition (CVPR)},
  year      = {2025}
}

@inproceedings{hao2025research,
  author    = {Ruiyang Hao and others},
  title     = {Research Challenges and Progress in the End-to-End {V2X} Cooperative Autonomous Driving Competition},
  booktitle = {Proceedings of the IEEE/CVF International Conference on Computer Vision (ICCV)},
  year      = {2025}
}

@inproceedings{mahajan2019maven,
  title     = {{MAVEN}: Multi-Agent Variational Exploration},
  author    = {Mahajan, Anuj and Rashid, Tabish and Samvelyan, Mikayel and Whiteson, Shimon},
  booktitle = {Advances in Neural Information Processing Systems},
  volume    = {32},
  year      = {2019}
}

@inproceedings{zhang2025planning,
  title     = {Planning with Multi-Constraints via Collaborative Language Agents},
  author    = {Zhang, Cong and others},
  booktitle = {Proceedings of the 31st International Conference on Computational Linguistics},
  year      = {2025}
}

@article{wang2002modeling,
  title     = {Modeling Constraint-Based Negotiating Agents},
  author    = {Wang, Huaiqing and Liao, Stephen and Liao, Lejian},
  journal   = {Decision Support Systems},
  volume    = {33},
  number    = {2},
  pages     = {201--217},
  year      = {2002}
}

@inproceedings{qin2025robofactory,
  title     = {RoboFactory: Exploring Embodied Agent Collaboration with Compositional Constraints},
  author    = {Qin, Yiran and others},
  booktitle = {Proceedings of the IEEE/CVF International Conference on Computer Vision},
  year      = {2025}
}

@article{bao2022recent,
  title     = {Recent Advances on Cooperative Control of Heterogeneous Multi-Agent Systems Subject to Constraints: A Survey},
  author    = {Bao, Guangyan and Ma, Lifeng and Yi, Xiaojian},
  journal   = {Systems Science \& Control Engineering},
  volume    = {10},
  number    = {1},
  pages     = {539--551},
  year      = {2022}
}

@misc{nauman2024overestimation,
  title        = {Overestimation, Overfitting, and Plasticity in Actor-Critic: {The} Bitter Lesson of Reinforcement Learning},
  author       = {Nauman, Michal and others},
  year         = {2024},
  eprint       = {2403.00514},
  archiveprefix = {arXiv}
}

\clearpage
\newpage

\appendix                    
\clearpage

\section{Limitations and Future Work}
\label{sec:limitations}

While the proposed Episodic Memory Temporal Consistency (EMTC) framework demonstrates significant improvements in sample efficiency and asymptotic performance across challenging cooperative MARL benchmarks, it is not without limitations. We explicitly acknowledge the following aspects, which present exciting avenues for future research:

\paragraph{Heuristic Annealing of Hyperparameters.} 
In the Temporal Consistency Gating Mechanism (TCGM), the temperature parameter $\tau$ dictates the strictness of the consistency filter. Currently, EMTC adopts a linear annealing schedule for $\tau$ (as detailed in Appendix \ref{Parametric Study on TCGM}) to balance early-stage exploration and late-stage exploitation. While empirically effective, this fixed-step annealing strategy ($\mathcal{T}_{decay}$) is largely heuristic and may require environment-specific tuning. Developing an adaptive scheduling mechanism, potentially based on the rolling variance of the temporal consistency error $\Delta_{\text{mem}}$, would make the framework more robust and plug-and-play.
\paragraph{Theoretical Gap in Lipschitz Continuity.} 
As established in Proposition \ref{prop:tcgm} and Equation \ref{eq:gap}, our theoretical guarantees rely on the assumption that the true optimal value function $V^*$ is $L$-Lipschitz continuous with respect to the learned latent space. While our TCSE module actively enforces a well-behaved and semantically clustered latent space to support this assumption, standard deep neural networks are not inherently Lipschitz bounded. In highly stochastic environments with sparse, discrete rewards, local discontinuities may still occur. Future work could explore incorporating explicit Lipschitz regularization techniques into the semantic embedder.

\section{Related Works}
\label{app:related}
This section reviews related work on incentive generation for exploration, as well as on constraints and overestimation regulation in multi-agent reinforcement learning.

\subsection{Incentive for Multi-Agent Exploration}
Balancing exploration and exploitation in MARL is notoriously difficult due to the combinatorial explosion of the joint action space. To address this, various intrinsic motivation and exploration incentive mechanisms have been proposed. For instance, MAVEN (\cite{mahajan2019maven}) and CDS (\cite{li2021celebrating}) leverage mutual information to encourage diverse and individualized behaviors. EMC (\cite{zheng2021episodic}) introduces curiosity-driven exploration by utilizing prediction errors of individual Q-values as intrinsic rewards. Beyond parametric intrinsic rewards, episodic control (\cite{blundell2016model, pritzel2017neural}) has been integrated into MARL to directly exploit high-return historical trajectories for sample-efficient learning. EMDQN (\cite{lin2018episodic}) utilized episodic memory to accelerate conventional Q-learning, an approach later extended to cooperative MARL frameworks. Recently, EMU (\cite{na2024efficient}) proposed explicit episodic incentives based on trajectory desirability to rapidly guide exploration toward successful states. However, these conventional frameworks typically inject episodic incentives indiscriminately. By relying on coarse, episode-level desirability metrics, they neglect the underlying temporal causality and semantic consistency of individual state transitions, frequently leaving agents vulnerable to suboptimal entrapment.

\subsection{Constraints and Overestimation Regulation in MARL}
Incorporating constraints into multi-agent reinforcement learning is essential for safe, robust, and practical coordination. 
A prominent line of work formulates the problem as a constrained cooperative control problem, where agents must optimize a shared objective while strictly satisfying operational or safety restrictions, as surveyed by (\cite{bao2022recent}). 
Early approaches modelled constraints within negotiation-based agent interactions (\cite{wang2002modeling}), while more recent works have introduced compositional constraints for embodied collaboration (\cite{qin2025robofactory}) and planning with multiple constraints via collaborative language agents (\cite{zhang2025planning}).

Beyond environmental safety, another critical algorithmic constraint in MARL lies in regulating Q-value estimates. 
Due to decentralized execution and the max operator over vast joint action spaces, value-based MARL is inherently susceptible to severe overestimation bias. 
Recent analysis has identified overestimation, together with overfitting and plasticity loss, as a principal failure mode in modern actor‑critic methods (\cite{nauman2024overestimation}), motivating explicit constraints on the learning dynamics.

\section{Mathematical Proof}
\label{appendix:proofs}

\subsection{Proof of Proposition \ref{prop:tcgm}}
\label{proof:t1}

To formally establish the reliability of the temporal consistency error $\Delta_{\text{mem}}$ as a gating signal, we must evaluate its relationship with the true transition suboptimality while accounting for representation and storage inaccuracies. 

\textbf{Error Decomposition of the Episodic Target.} During the learning process, for any encountered transition $(s, a, r, s')$, the algorithm retrieves the highest empirical return $\mathcal{H}$ from the $\delta$-nearest neighbors in the latent space, defined as $\hat{V}_E(s) \doteq \mathcal{H}(\hat{s})$ and $\hat{V}_E(s') \doteq \mathcal{H}(\hat{s}')$. The total approximation error at state $s$ can be decoupled using the triangle inequality:
\begin{equation}
    |\hat{V}_E(s) - V^*(s)| = |\mathcal{H}(\hat{s}) - V^*(s)| \le |\mathcal{H}(\hat{s}) - V^*(\hat{s})| + |V^*(\hat{s}) - V^*(s)|.
\end{equation}
This error inherently consists of two components: the storage error $\epsilon_{\text{store}}(\hat{s}) \doteq |\mathcal{H}(\hat{s}) - V^*(\hat{s})|$ and the representation error $\epsilon_{\text{rep}}(s) \doteq |V^*(\hat{s}) - V^*(s)|$. Defining the global maximum over the state space as $\epsilon_{\text{store}}$ and $\epsilon_{\text{rep}}$, the total approximation error is strictly bounded by $\epsilon_{\text{rep}} + \epsilon_{\text{store}}$.

\textbf{Bounding the Temporal Consistency Error.} The temporal consistency error utilized by our gating mechanism is calculated as $\Delta_{\text{mem}} = |r + \gamma \hat{V}_E(s') - \hat{V}_E(s)|$. To evaluate its reliability, we inject the optimal value function $V^*$ by adding and subtracting $V^*(s)$ and $\gamma V^*(s')$:
\begin{equation}
    \Delta_{\text{mem}} = \left| (r + \gamma V^*(s') - V^*(s)) + \gamma(\hat{V}_E(s') - V^*(s')) - (\hat{V}_E(s) - V^*(s)) \right|.
\end{equation}
By applying the triangle inequality, we isolate the true underlying optimality gap of the executed transition, denoted as $\Delta_{\text{opt}} \doteq |r + \gamma V^*(s') - V^*(s)|$:
\begin{equation}
    \Delta_{\text{mem}} \le \Delta_{\text{opt}} + \gamma |\hat{V}_E(s') - V^*(s')| + |\hat{V}_E(s) - V^*(s)|.
\end{equation}
Substituting the maximum approximation error bounds derived above, we obtain:
\begin{equation}
    \Delta_{\text{mem}} \le \Delta_{\text{opt}} + (1+\gamma)(\epsilon_{\text{rep}} + \epsilon_{\text{store}}).
\end{equation}
For simplicity, we define the aggregated noise induced by representation and storage as $\alpha \doteq (1+\gamma)(\epsilon_{\text{rep}} + \epsilon_{\text{store}})$, allowing us to concisely express the upper bound as:
\begin{equation} \label{eq:delta_bound}
    \Delta_{\text{mem}} \le \Delta_{\text{opt}} + \alpha.
\end{equation}

\textbf{The Role of Temporally Consistent Semantic Embedding.} Equation \eqref{eq:delta_bound} reveals the precise operational condition required for a safe gating mechanism. If the representation collapses, the retrieved state $\hat{s}$ severely mismatches the true semantic state $s$, causing $\epsilon_{\text{rep}}$ and consequently $\alpha$ to explode. Under such representation failures, a genuinely optimal trajectory ($\Delta_{\text{opt}} \approx 0$) might still yield a massive $\Delta_{\text{mem}}$, causing the gate to erroneously penalize high-quality experiences. Crucially, this failure mode is explicitly prevented by the Temporally Consistent Semantic Embedder (TCSE). Assuming $V^*$ is $L$-Lipschitz continuous with respect to the latent space, the representation error is theoretically bounded by the retrieval radius $\delta$:
\begin{equation}
    \epsilon_{\text{rep}} \le L\|f_\phi(\hat{s}) - f_\phi(s)\|_2 \le L\delta.
\label{eq:gap}
\end{equation}
The dual-objective optimization of TCSE actively prevents representation collapse and preserves semantic discriminability, ensuring that the Lipschitz constant $L$ remains well-behaved and that the $\delta$-radius retrieval successfully identifies true semantic equivalents. Consequently, TCSE strictly limits $\epsilon_{\text{rep}}$, while periodic memory updates incrementally minimize $\epsilon_{\text{store}}$.

\textbf{Reliability of the Gating Mechanism.} This algorithmic synergy guarantees that the joint noise $\alpha$ is systematically minimized and bounded at a low magnitude throughout training. When encountering a misleading pseudo-successful trajectory, the true suboptimality is large ($\Delta_{\text{opt}} \gg \alpha$), ensuring that $\Delta_{\text{mem}}$ accurately tracks $\Delta_{\text{opt}}$ and safely activates the penalty. Conversely, for a genuinely high-quality trajectory where $\Delta_{\text{opt}} \approx 0$, the bounded nature of $\alpha$ ensures that $\Delta_{\text{mem}}$ remains sufficiently small ($\Delta_{\text{mem}} \le \alpha \approx 0$). This yields a gating coefficient $\beta \to 1$, preserving the optimal reward injection. Thus, our framework rigorously proves that effective incentive gating is strictly contingent upon robust semantic representation. \hfill $\blacksquare$

\subsection{Proof of Proposition \ref{prop:asymp}}
\label{proof:t2}

Recall from Eq.~\eqref{eq:gradient} that the gradient signal derived from the consistency-aware episodic reward is given by:
\begin{equation}
    \nabla_\theta \mathcal{L}^\beta_\theta = -2 \nabla_\theta Q_{tot}(s, a; \theta) \left( \Delta_{\text{TD}} + \tilde{r}_p(s, a) \right),
    \label{eq:proof_grad_start}
\end{equation}
where the standard one-step TD error is $\Delta_{\text{TD}} = r + \gamma \max_{a'} Q_{\theta^-}(s', a') - Q_{tot}(s, a; \theta)$, and the gated episodic incentive is defined as:
\begin{equation}
    \tilde{r}_p(s, a) = \beta(\Delta_{\text{mem}}) \cdot \gamma \frac{N_\xi(s')}{N_{\text{call}}(s')} \left( \mathcal{H}(s') - \max_{a'} Q_{\theta^-}(s', a') \right).
    \label{eq:proof_rp}
\end{equation}

\textbf{Asymptotic Behaviour Setup.} To analyze the asymptotic behavior, consider the training process as the joint policy converges to the optimal policy ($\pi_\theta \to \pi^*$) and the memory buffer matures.

\textbf{Convergence of the Gating Coefficient.} According to \textbf{Proposition \ref{prop:tcgm}}, as the episodic operator continuously updates under adequate exploration, the memory transitions inherently satisfy the Bellman optimality equation, causing the temporal consistency error to vanish ($\Delta_{\text{mem}} \to 0$). Based on the definition of our Gaussian-style gating function in Eq.~\eqref{eq:beta_gate}, this ensures that the consistency coefficient asymptotically approaches exactly 1:
\begin{equation}
    \lim_{\mathcal{H} \to V^*} \beta(\Delta_{\text{mem}}) = 1.
    \label{eq:proof_limit_beta}
\end{equation}

\textbf{Convergence of the Desirability Ratio.} When the actions follow the optimal policy $a \sim \pi^*$, every retrieved trajectory in $\mathcal{D}_E$ naturally leads to a desirable outcome. Thus, the desirability indicator $\xi(s') = 1$ consistently holds. As a result, the ratio of desirable transitions to total visits converges to 1:
\begin{equation}
    \lim_{\pi_\theta \to \pi^*} \frac{N_\xi(s')}{N_{\text{call}}(s')} = 1.
    \label{eq:proof_limit_ratio}
\end{equation}

\textbf{Asymptotic Reward Recovery.} By substituting the limits from Eq.~\eqref{eq:proof_limit_beta} and Eq.~\eqref{eq:proof_limit_ratio} into Eq.~\eqref{eq:proof_rp}, the asymptotic behavior of the gated episodic reward becomes:
\begin{equation}
    \lim_{\pi_\theta \to \pi^*, \mathcal{H} \to V^*} \tilde{r}_p(s, a) = \gamma \left( V^*(s') - \max_{a'} Q_{\theta^-}(s', a') \right).
    \label{eq:proof_limit_rp}
\end{equation}

\textbf{Gradient Consistency Proof.} Finally, we substitute this converged incentive back into the overall gradient equation (Eq.~\eqref{eq:proof_grad_start}):
\begin{equation}
\label{eq:proof_grad_final}
\begin{aligned}
    \lim_{\pi_\theta \to \pi^*, \mathcal{H} \to V^*} \nabla_\theta \mathcal{L}^\beta_\theta 
    &= -2 \nabla_\theta Q_{tot}(s, a; \theta) \left[ \Delta_{\text{TD}} + \lim_{\pi_\theta \to \pi^*, \mathcal{H} \to V^*} \tilde{r}_p(s, a) \right] \\
    &= -2 \nabla_\theta Q_{tot}(s, a; \theta) \bigg[ r + \gamma \max_{a'} Q_{\theta^-}(s', a') - Q_{tot}(s, a; \theta) \\
    &\qquad \qquad \qquad \qquad \quad + \gamma V^*(s') - \gamma \max_{a'} Q_{\theta^-}(s', a') \bigg] \\
    &= -2 \nabla_\theta Q_{tot}(s, a; \theta) \left[ r + \gamma V^*(s') - Q_{tot}(s, a; \theta) \right] \\
    &= \nabla_\theta \mathcal{L}^*_\theta.
\end{aligned}
\end{equation}

\textbf{Unbiasedness Conclusion.} Therefore, as the policy reaches optimality, the consistency-aware gradient $\nabla_\theta \mathcal{L}^\beta_\theta$ perfectly cancels out the estimation bias introduced by the target network $\max_{a'} Q_{\theta^-}(s', a')$, and gracefully degenerates to the unbiased optimal gradient signal $\nabla_\theta \mathcal{L}^*_\theta$ guided directly by $V^*(s')$. This proves that the gating mechanism preserves the original optimization fixed point without introducing any asymptotic bias. \hfill $\blacksquare$

\section{Implementation and Experiment Details}   
\subsection{Supplementary Information on the Experimental Environment}
\label{app:environment}
In this section, we detail the environmental configurations, including the dimensions of the state and action spaces for both the StarCraft Multi-Agent Challenge (SMAC) \cite{samvelyan2019starcraft} and Google Research Football (GRF) \cite{kurach2020google} benchmarks.

\subsubsection{StarCraft Multi-Agent Challenge (SMAC)}
 
 In SMAC, the global state encompasses the exact coordinates of all agents along with the specific features of both allied and enemy units. The discrete action space includes movement directions and targeted attacks, meaning its size naturally scales with the number of enemies in a given scenario. As summarized in Table~\ref{tab:smac_dims}, the exact dimensions of the state and action spaces vary across different maps. We utilize the default shaped reward structure for all scenarios, which grants positive signals for dealing damage to enemies and an additional bonus for winning the battle. 

\begin{table}[htbp]
\centering
\caption{Dimension of the state space and the action space of SMAC}
\label{tab:smac_dims}
\begin{tabular}{lcc}
\toprule
\textbf{Task} & \textbf{Dimension of state space} & \textbf{Dimension of action space} \\
\midrule
1c3s5z        & 270  & 15 \\
3s5z          & 216  & 14 \\
3s\_vs\_5z    & 68   & 11 \\
5m\_vs\_6m    & 98   & 12 \\
MMM2          & 322  & 18 \\
6h\_vs\_8z    & 140  & 14 \\
3s5z\_vs\_3s6z& 230  & 15 \\
corridor      & 282  & 30 \\
\bottomrule
\end{tabular}
\end{table}

\subsubsection{Google Research Football (GRF)}

 In the GRF environment, the global state provides comprehensive match information, including player coordinates, orientations, and ball possession. While the state dimension varies by scenario, the action space remains strictly uniform across all tasks. Each agent can execute a predefined set of actions, including moving, various kicking techniques, sprinting, intercepting, and dribbling. Table~\ref{tab:grf_dims} summarizes these dimensionalities. Regarding the reward function, GRF offers both dense and sparse modes. In our experiments, we strictly adopt the challenging sparse reward setting, where agents receive a reward of $+1$ for scoring a goal and $-1$ for conceding one.

\begin{table}[htbp]
\centering
\caption{Dimension of the state space and the action space of GRF}
\label{tab:grf_dims}
\begin{tabular}{lcc}
\toprule
\textbf{Task} & \textbf{Dimension of state space} & \textbf{Dimension of action space} \\
\midrule
3\_vs\_1WK & 26 & 19 \\
CA\_easy   & 30 & 19 \\
CA\_hard   & 34 & 19 \\
\bottomrule
\end{tabular}
\end{table}

\subsection{Experiment Details}
\label{app:experiment details}
We utilize PyMARL \cite{samvelyan2019starcraft} to execute all of the baseline algorithms with their
open-source codes, and the same hyperparameters are used for experiments if they are presented
either in uploaded codes or in their manuscripts.
All SMAC experiments were conducted on StarCraft II version 4.10.0 in a Linux environment.

For Google Research Football task, we use the environmental code provided by \cite{wang2021rode}. In the experiments, we consider three official scenarios such as academy\_3\_vs\_1\_with\_keeper
(3\_vs\_1WK), academy\_counterattack\_easy (CA\_easy), and academy\_counterattack\_hard (CA\_hard).

Win-rate is computed with 80 samples: 16 episodes for each training random seed, and 5 different random seeds unless denoted otherwise. Both the mean and the variance of the performance are presented for all the figures to show their overall performance according to different seeds. EMTC set task-dependent $\delta$ values as presented in Table~\ref{tab:emtc_delta}.
For other hyperparameters introduced by EMTC, the same values presented in Table \ref{tab:emtc_hyperparams} are used throughout all tasks.
For EMTC (QPLEX) in \textit{corridor}, $\delta = 1.8 \times 10^{-5}$ is used instead of $\delta = 1.8 \times 10^{-3}$. Appendix \ref{app:tau} presents the discussion about $\tau$.

\begin{table}[ht]
  \centering
  \caption{Task-dependent Hyperparameter of EMTC.}
  \label{tab:emtc_delta}
  \begin{tabular}{ll}
    \toprule
    Category & $\delta$ \\
    \midrule
    easy/hard SMAC maps & $1.8 \times 10^{-5}$\\
    super hard SMAC maps & $1.8 \times 10^{-3}$ \\
    GRF & $1.8 \times 10^{-3}$ \\
    \bottomrule
  \end{tabular}
\end{table}

\begin{table}[ht]
  \centering
  \caption{Other EMTC Hyperparameters }
  \label{tab:emtc_hyperparams}
  \begin{tabular}{ll}
    \toprule
    Configuration & Value \\
    \midrule
    a scale factor of reconstruction loss, $\lambda_{\text{rcon}}$ & 0.1 \\
    a scale factor of contrastive loss, $\lambda_{\text{cl}}$ & 0.1 \\
    a scale factor of contrastive learning, $\tau_{cl}$ & 0.1 \\
    update interval, $t_{\text{emb}}$ & 1K \\
    training samples, $N$ & 102.4K \\
    batch size of training, $B$ & 1024 \\
    episodic latent dimension, $\dim(x)$ & 4 \\
    episodic memory capacity & 1M \\
    a scale factor, $\lambda$ (for conventional episodic control only) & 0.1 \\
    \bottomrule
  \end{tabular}
\end{table}

\subsection{Infrastructure and Computational Overhead}
\label{appendix:infrastructure}

All experiments for the SMAC and GRF environments were conducted on a computing cluster equipped with NVIDIA A100 Tensor Core GPUs and high-performance CPUs.

In terms of computational overhead, \sysname{} is highly efficient and comparable to the state-of-the-art baseline, EMU. For the most computationally demanding super-hard scenario in SMAC, such as \textit{corridor}, the total training process takes approximately 15 hours on a single A100 GPU. 

Although \sysname{} introduces an additional contrastive learning objective ($\mathcal{L}_{cl}$) and a Temporal Consistency Gating Mechanism ($\beta$), they do not introduce significant computational bottlenecks. Specifically:
\begin{itemize}
    \item \textbf{Periodic Embedder Update:} Similar to EMU, the semantic embedder ($f_\phi$ and $f_\psi$) is updated periodically rather than at every timestep. Benefiting from the high memory bandwidth and Tensor Cores of the A100 GPU, the forward-backward pass for the dual-objective learning (time-augmented reconstruction and contrastive InfoNCE) and the subsequent re-projection of the episodic buffer $\mathcal{D}_E$ are extremely fast, taking less than 2 seconds per update cycle.
    \item \textbf{Gating Mechanism Calculation:} The computation of the temporal consistency error $\Delta_{\text{mem}}$ and the gating coefficient $\beta(\Delta_{\text{mem}})$ is essentially an $\mathcal{O}(1)$ scalar operation executed during the standard one-step TD inference. It naturally integrates into the vectorized Bellman updates on the GPU, incurring virtually zero extra wall-clock time.
\end{itemize}

To further verify that our embedder TCSE introduces no additional inference latency compared to the conventional dCAE encoder (as used in EMU), we benchmarked the per-step forward time of both architectures across representative state dimensions and batch sizes. As summarized in Table~\ref{tab:encoder_efficiency}, the TCSE encoder achieves virtually identical computational efficiency with an average speedup of $1.01\times$ over the dCAE baseline and a memory footprint ratio of $0.99\times$. This result confirms that the enhanced representation quality of TCSE comes at essentially zero extra cost at inference time.

\begin{table}[t]
\centering
\caption{Per-step Forward Time Comparison.}
\label{tab:encoder_efficiency}
\begin{tabular}{cccccc}
\toprule
\textbf{State Dim} & \textbf{Batch Size} & \textbf{Seq Len} & \textbf{dCAE Time (ms)} & \textbf{TCSE Time (ms)} & \textbf{Speedup} \\
\midrule
128 & 1  & 1   & 0.100 & 0.090 & 0.90$\times$ \\
128 & 8  & 10  & 0.110 & 0.110 & 1.00$\times$ \\
128 & 32 & 50  & 0.090 & 0.099 & 1.10$\times$ \\
128 & 64 & 100 & 0.110 & 0.080 & 0.73$\times$ \\
128 & 64 & 1   & 0.105 & 0.100 & 0.95$\times$ \\
\midrule
512 & 1  & 10  & 0.135 & 0.110 & 0.81$\times$ \\
512 & 8  & 50  & 0.138 & 0.116 & 0.84$\times$ \\
512 & 32 & 10  & 0.106 & 0.090 & 0.85$\times$ \\
512 & 64 & 1   & 0.111 & 0.130 & 1.17$\times$ \\
512 & 64 & 50  & 0.070 & 0.081 & 1.16$\times$ \\
\midrule
\multicolumn{3}{c}{\textbf{Overall Average}} & --- & --- & \textbf{1.01$\times$} \\
\bottomrule
\end{tabular}
\end{table}

Consequently, the extra computational time required for our temporally consistent memory module is strictly marginal, establishing \sysname{} as a highly practical algorithm for complex MARL tasks.

\section{Further Experiment Result} 

\subsection{Indicator Explanation}
\label{app:uw}
To comprehensively evaluate the performance of our proposed method, relying solely on the final test win-rate at the end of training is insufficient, as it fails to capture the learning efficiency. Therefore, following the evaluation protocol introduced in EMU \cite{na2024efficient}, we adopt a composite evaluation metric denoted as the Normalized Overall Win-Rate, $\bar{\mu}_w(t)$. 

\begin{figure}[htbp]
    \centering
    \includegraphics[width=0.5\linewidth]{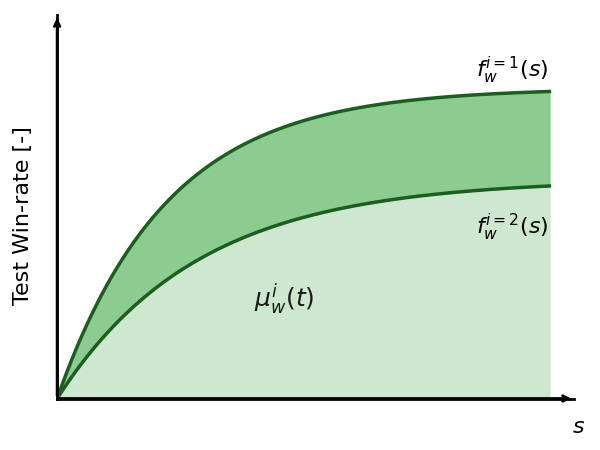}
    \caption{Visual illustration of the Normalized Overall Win-Rate $\bar{\mu}_w(t)$ and $s$ is denoted as the training time. The shaded area corresponds to the cumulative test win-rate $\int_{0}^{t} f_w^i(s) ds$. By calculating this area under the curve, $\bar{\mu}_w(t)$ evaluates not only the final asymptotic win-rate but also the learning efficiency, rewarding algorithms that converge faster to optimal policies.}
    \label{fig:muw_illustration}
\end{figure}

Let $f_w^i(s)$ define the test win-rate achieved by the agent at training step $s$ under the $i$-th random seed. As illustrated in Figure~\ref{fig:muw_illustration}, the cumulative performance of a single training run up to time $t$ can be geometrically interpreted as the area under its learning curve, calculated by the integral $\mu_w^i(t) = \int_{0}^{t} f_w^i(s) ds$. To aggregate this performance across $n$ independent runs and bound the metric within the range $[0, 1]$, we average these integrals and divide by the theoretical maximum area $\mu_{max} = t$. The formal calculation is defined as:

\begin{equation}
\bar{\mu}_w(t) = \frac{1}{t \cdot n} \sum_{i=1}^{n} \int_{0}^{t} f_w^i(s) ds 
\end{equation}

\label{app:further_experiments}

\subsection{Detailed Quantitative Results}
\label{app:quantitative_results}

To complement the learning curves presented in Section \ref{sec:exp_grf}, we provide the detailed quantitative results of all evaluated algorithms across the StarCraft Multi-Agent Challenge (SMAC) and Google Research Football (GRF) environments. 

Table~\ref{tab:smac_results} and Table~\ref{tab:grf_results} summarize the median of the final test win rates across multiple independent runs with different random seeds. The results clearly demonstrate that \sysname{} consistently achieves the highest final performance.

\begin{table*}[htbp]
\centering
\caption{Final test win rates (\%) on six SMAC scenarios. Results are shown as mean $\pm$ standard deviation across multiple independent runs. The best performances are highlighted in \textbf{bold}.}
\label{tab:smac_results}
\resizebox{\textwidth}{!}{
\begin{tabular}{lcccccc}
\toprule
\textbf{Algorithm} & \textbf{1c3s5z} & \textbf{3s\_vs\_5z} & \textbf{MMM2} & \textbf{3s5z\_vs\_3s6z} & \textbf{corridor} & \textbf{6h\_vs\_8z} \\
\midrule
\textbf{EMTC (CDS)}   & $\mathbf{100.0 \pm 0.0}$ & $\mathbf{100.0 \pm 0.0}$ & $\mathbf{100.0 \pm 0.0}$ & $\mathbf{98.4 \pm 1.5}$  & $\mathbf{99.2 \pm 0.5}$  & $\mathbf{89.4 \pm 3.2}$ \\
\textbf{EMTC (QPLEX)} & $\mathbf{100.0 \pm 0.0}$ & $\mathbf{100.0 \pm 0.0}$ & $99.3 \pm 0.3$            & $94.0 \pm 2.1$           & $97.4 \pm 1.7$           & $75.0 \pm 4.8$          \\
\midrule
EMU (CDS)             & $\mathbf{100.0 \pm 0.0}$ & $92.1 \pm 2.7$           & $99.3 \pm 0.5$           & $92.5 \pm 2.5$           & $84.0 \pm 3.8$           & $65.2 \pm 5.6$          \\
EMU (QPLEX)           & $\mathbf{100.0 \pm 0.0}$ & $90.3 \pm 2.9$           & $98.6 \pm 1.3$           & $78.2 \pm 5.2$           & $73.4 \pm 5.0$           & $60.5 \pm 6.0$          \\
\midrule
CDS                   & $\mathbf{100.0 \pm 0.0}$ & $85.0 \pm 4.8$           & $95.1 \pm 3.4$           & $91.1 \pm 3.0$           & $81.5 \pm 4.3$           & $29.4 \pm 9.8$          \\
QPLEX                 & $96.8 \pm 2.1$           & $50.6 \pm 8.4$           & $96.9 \pm 2.7$           & $81.3 \pm 4.5$           & $75.0 \pm 5.0$           & $7.8 \pm 2.8$           \\
QMIX                  & $97.6 \pm 2.0$           & $70.4 \pm 5.6$           & $92.4 \pm 3.7$           & $67.2 \pm 6.0$           & $37.6 \pm 9.8$           & $12.8 \pm 4.4$          \\
\bottomrule
\end{tabular}
}
\end{table*}

\begin{table}[htbp]
\centering
\caption{Final test win rates (\%) on three Google Research Football (GRF) scenarios. Results are shown as mean $\pm$ standard deviation across multiple independent runs.}
\label{tab:grf_results}
\begin{tabular}{lccc}
\toprule
\textbf{Algorithm} & \textbf{3\_vs\_1WK} & \textbf{CA\_easy} & \textbf{CA\_hard} \\
\midrule
\textbf{EMTC (CDS)}   & $\mathbf{94.5 \pm 1.4}$ & $\mathbf{72.1 \pm 2.2}$ & $48.0 \pm 2.8$          \\
\textbf{EMTC (QPLEX)} & $91.3 \pm 1.3$          & $64.1 \pm 2.3$          & $\mathbf{62.6 \pm 2.5}$ \\
\midrule
EMU (CDS)             & $65.1 \pm 3.2$          & $50.4 \pm 3.5$          & $34.8 \pm 3.0$          \\
EMU (QPLEX)           & $50.2 \pm 4.6$          & $25.0 \pm 4.0$          & $37.9 \pm 3.4$          \\
\midrule
CDS                   & $30.3 \pm 7.1$          & $5.1 \pm 2.2$           & $12.7 \pm 4.0$          \\
QPLEX                 & $10.9 \pm 3.2$          & $10.2 \pm 3.1$          & $10.0 \pm 3.5$          \\
QMIX                  & $30.2 \pm 7.4$          & $17.0 \pm 5.7$          & $25.6 \pm 6.8$          \\
\bottomrule
\end{tabular}
\end{table}

\subsection{Ablation on TCSE without TCGM}
\label{app:ablation_emb}

In the experimental evaluation presented in Section \ref{sec:experiments}, we utilized the composite metric $\bar{\mu}_w(t)$ to investigate the embedding efficacy on the \texttt{3s\_vs\_5z} and \texttt{6h\_vs\_8z} maps. In this section, we extend this analysis to further isolate and rigorously evaluate the representation quality of different semantic embedders. Specifically, we focus on two distinct scenarios: the relatively simple map \texttt{3s\_vs\_5z} and the super-hard map \texttt{corridor}.

To ensure that the performance variations are strictly attributable to the embedding structures themselves, we completely deactivate the Temporal Consistency Gating Mechanism ($\beta$) for these evaluations. We compare three distinct embedding designs: Random Projection, deterministic Conditional Autoencoder (dCAE), and our proposed Temporally Consistent Semantic Embedder (TCSE).

\begin{table}[htbp]
\centering
\caption{$\bar{\mu}_w$ according to $\delta$ and design choice of embedding function on \textbf{easy} SMAC map, \texttt{3s\_vs\_5z}. }
\label{tab:embed_3s5z}
\renewcommand{\arraystretch}{1.2}
\setlength{\tabcolsep}{5pt}
\begin{tabular}{c ccc ccc ccc}
\toprule
\multirow{2}{*}[0pt]{\begin{tabular}[c]{@{}c@{}}Steps{[}mil{]}\end{tabular}} & \multicolumn{3}{c}{0.8} & \multicolumn{3}{c}{1.6} & \multicolumn{3}{c}{2.4} \\ 
\cmidrule(lr){2-4} \cmidrule(lr){5-7} \cmidrule(lr){8-10}
$\delta$ & random & dCAE & TCSE & random & dCAE & TCSE & random & dCAE & TCSE \\
\midrule
1.8e-7 & 0.087 & 0.102 & \textbf{0.119} & 0.211 & 0.283 & \textbf{0.326} & 0.401 & 0.501 & \textbf{0.550} \\
1.8e-5 & 0.083 & 0.112 & \textbf{0.120} & 0.225 & 0.298 & \textbf{0.315} & 0.424 & 0.527 & \textbf{0.541} \\
1.8e-3 & 0.090 & 0.091 & \textbf{0.102} & 0.237 & 0.244 & \textbf{0.280} & 0.446 & 0.442 & \textbf{0.514} \\
1.8e-2 & 0.075 & \textbf{0.102} & 0.098 & 0.201 & \textbf{0.265} & 0.252 & 0.384 & \textbf{0.470} & 0.458 \\
\bottomrule
\end{tabular}
\end{table}

\begin{table}[htbp]
\centering
\caption{$\bar{\mu}_w$ according to $\delta$ and design choice of embedding function on \textbf{super-hard} SMAC map, \texttt{corridor}. }
\label{tab:embed_corridor}
\renewcommand{\arraystretch}{1.2}
\setlength{\tabcolsep}{5pt}
\begin{tabular}{c ccc ccc ccc}
\toprule
\multirow{2}{*}[0pt]{\begin{tabular}[c]{@{}c@{}}Steps{[}mil{]}\end{tabular}} & \multicolumn{3}{c}{0.8} & \multicolumn{3}{c}{1.6} & \multicolumn{3}{c}{2.4} \\ 
\cmidrule(lr){2-4} \cmidrule(lr){5-7} \cmidrule(lr){8-10}
$\delta$ & random & dCAE & TCSE & random & dCAE & TCSE & random & dCAE & TCSE \\
\midrule
1.8e-7 & 0.003 & 0.025 & \textbf{0.033} & 0.038 & 0.145 & \textbf{0.171} & 0.110 & 0.315 & \textbf{0.350} \\
1.8e-5 & 0.002 & 0.021 & \textbf{0.032} & 0.024 & 0.136 & \textbf{0.180} & 0.089 & 0.321 & \textbf{0.362} \\
1.8e-3 & 0.006 & 0.037 & \textbf{0.044} & 0.047 & 0.152 & \textbf{0.187} & 0.105 & 0.325 & \textbf{0.377} \\
1.8e-2 & 0.001 & 0.018 & \textbf{0.031} & 0.019 & 0.125 & \textbf{0.150} & 0.065 & 0.302 & \textbf{0.338} \\
\bottomrule
\end{tabular}
\end{table}

Tables \ref{tab:embed_3s5z} and \ref{tab:embed_corridor} present the $\bar{\mu}_w(t)$ values recorded at critical training milestones: 0.8M, 1.6M, and 2.4M timesteps, across various search radius thresholds $\delta$. To ensure a fair and rigorous comparison, the candidate values for $\delta$ are directly adopted from the evaluation protocol established in the EMU baseline(~\cite{na2024efficient}). 

As demonstrated in the tables, by eliminating the regulatory effect of $\beta$, the inherent representation collapse issues of basic embedders become more pronounced. In contrast, the TCSE consistently maintains superior $\bar{\mu}_w(t)$ scores across both easy and super-hard scenarios. This confirms that the synergy of contrastive learning and time-augmented reconstruction intrinsically provides a much more robust and semantically meaningful memory retrieval space, even without downstream incentive regulation.

\subsection{Dynamics of Temporal Consistency Error and Consistency Coefficient}
\label{app:delta_beta_dynamics}

To empirically validate the theoretical properties of our gating mechanism in Section \ref{subsec:gating}, we track the evolution of the Temporal Consistency Error ($\Delta_{\text{mem}}$) and the Consistency Coefficient ($\beta$) during training. Specifically, we sampled the average $\bar{\Delta}_{\text{mem}}$ and $\bar{\beta}$ from the episodic buffer $\mathcal{D}_E$ every 10,000 environment steps on the SMAC (\texttt{3s\_vs\_5z}) and GRF (\texttt{3\_vs\_1WK}) environments.

\begin{figure}[htbp]
    \centering
    \begin{minipage}[t]{0.48\textwidth}
        \centering
        \includegraphics[width=\linewidth]{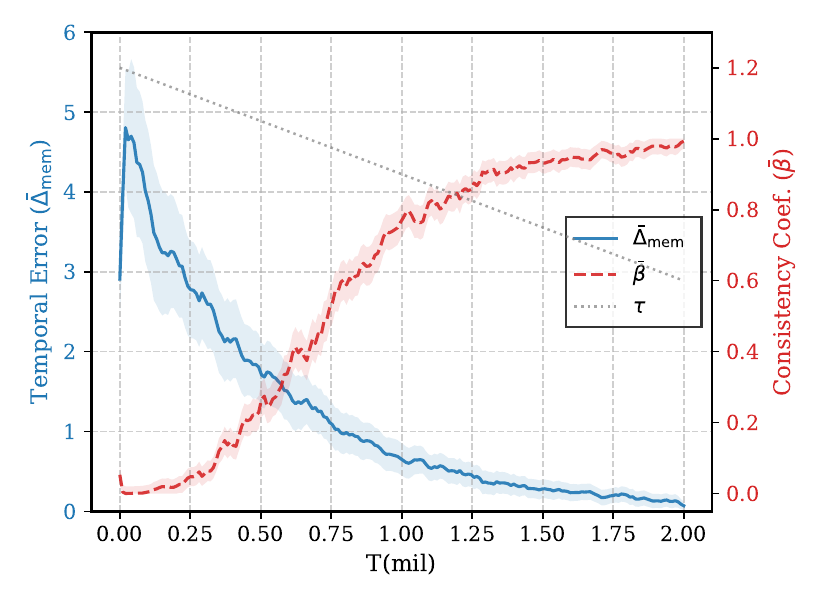} 
        \vspace{1mm}
        
        \small (a) \texttt{3s\_vs\_5z}
    \end{minipage}%
    \hfill
    \begin{minipage}[t]{0.48\textwidth}
        \centering
        \includegraphics[width=\linewidth]{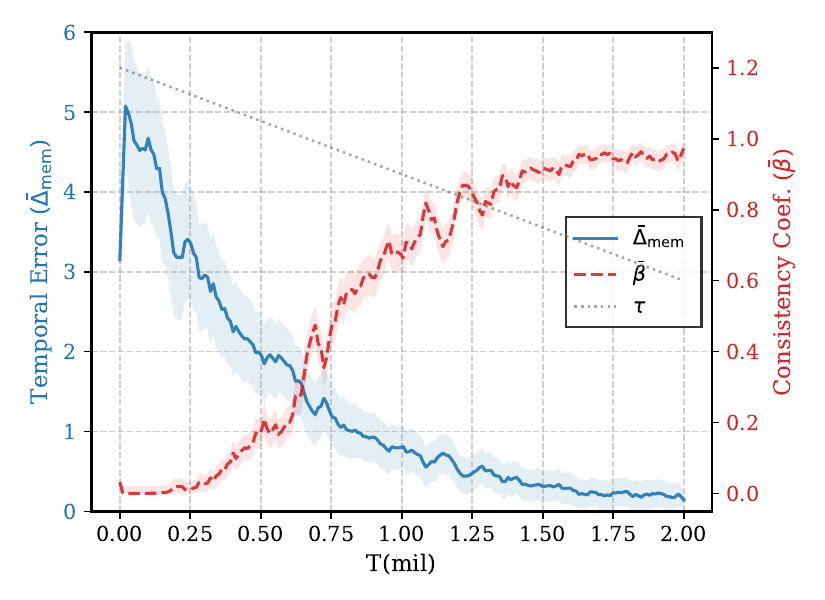} 
        \vspace{1mm}
        
        \small (b) \texttt{3\_vs\_1WK}
    \end{minipage}
    
    \vspace{2mm} 
    \caption{Evolution of the average Temporal Consistency Error ($\bar{\Delta}_{\text{mem}}$) and Consistency Coefficient ($\bar{\beta}$) alongside the annealed temperature $\tau$.}
    \label{fig:delta_beta_curve}
\end{figure}

As illustrated in Figure \ref{fig:delta_beta_curve}, the empirical results demonstrate a highly adaptive dynamic interplay among the temporal error ($\bar{\Delta}_{\text{mem}}$), the consistency coefficient ($\bar{\beta}$), and the temperature parameter ($\tau$). During the early stages of training, the episodic buffer predominantly contains disjointed exploratory trajectories, yielding a notably high $\bar{\Delta}_{\text{mem}}$. Driven by this large error, the gating mechanism aggressively suppresses $\bar{\beta}$ to near zero, effectively shielding the agent from misleading episodic incentives. As training progresses, high-quality and temporally coherent experiences accumulate. Consistent with our theoretical analysis, $\bar{\Delta}_{\text{mem}}$ drops rapidly and converges toward zero. Crucially, while $\tau$ is systematically annealed to enforce progressively stricter consistency bounds, the rapid decay of $\bar{\Delta}_{\text{mem}}$ outpaces this restriction, allowing $\bar{\beta}$ to naturally recover and approach $1.0$. This cross-domain evaluation verifies that the \sysname{} gating mechanism robustly penalizes pseudo-memories during initial exploration, while smoothly transitioning to deliver unattenuated, unbiased guidance as the joint policy approaches convergence.

\subsection{Parametric Study on TCSE}
\label{app:Parametric Study on TCSE}

\begin{figure}[htbp]
    \centering
    \begin{minipage}[t]{0.48\textwidth}
        \centering
        \includegraphics[width=\linewidth]{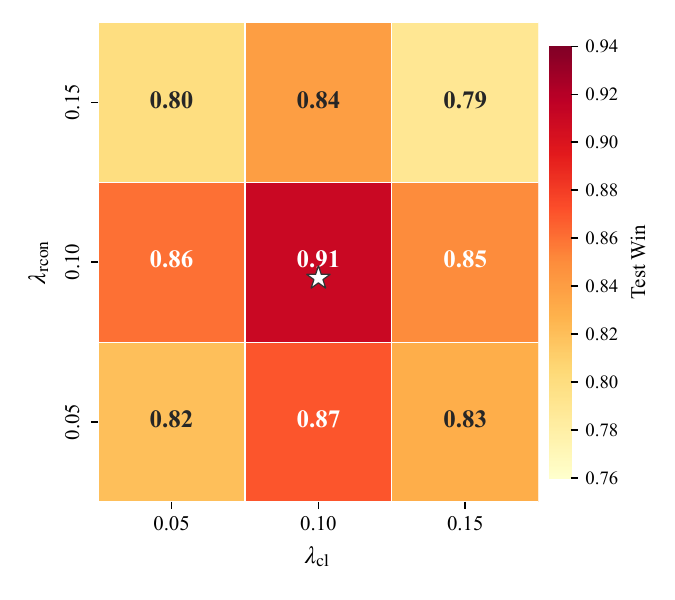} 
        \vspace{1mm}
        
        \small (a) \texttt{3\_vs\_1WK}
    \end{minipage}%
    \hfill
    \begin{minipage}[t]{0.48\textwidth}
        \centering
        \includegraphics[width=\linewidth]{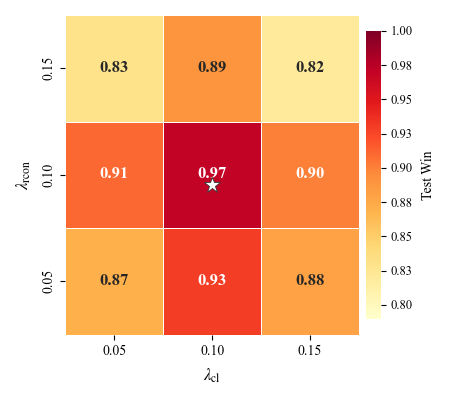} 
        \vspace{1mm}
        
        \small (b) \texttt{corridor}
    \end{minipage}
    
    \vspace{2mm} 
    \caption{Impact of $\lambda_{rcon}$ and $\lambda_{cl}$ on final test win-rates. The star ($\star$) indicates the optimal configuration.}
    \label{fig:lambda_heatmap}
\end{figure}

In this section, we conduct a comprehensive parametric study to evaluate the impact of two crucial hyperparameters in the Temporally Consistent Semantic Embedder (TCSE): the reconstruction loss weight ($\lambda_{rcon}$) and the contrastive learning loss weight ($\lambda_{cl}$). The primary objective of this experiment is to identify the optimal configuration that maximizes the embedder's representation quality by effectively balancing temporal-aware state feature extraction and semantic discriminability.

To thoroughly investigate this, we performed a grid search over $\lambda_{rcon}, \lambda_{cl} \in \{0.05, 0.10, 0.15\}$. The evaluation was carried out on two representative and challenging scenarios: the \texttt{3\_vs\_1WK} task from Google Research Football (GRF) and the super-hard \texttt{corridor} map from the StarCraft Multi-Agent Challenge (SMAC). As illustrated by the heatmaps in Figure \ref{fig:lambda_heatmap}, our \sysname{} framework consistently achieves peak performance across both domains when the parameters are set to $\lambda_{rcon} = 0.10$ and $\lambda_{cl} = 0.10$ (marked by the white star). 

\subsection{Data Augmentation Strategies for TCSE}
\label{app:augmentation}

The efficacy of contrastive learning relies heavily on the construction of meaningful positive pairs. In visual domains, standard image transformations (e.g., cropping, shifting) are widely applied. However, as recent studies highlight, directly applying spatial transformations to 1D multi-agent state vectors destroys fundamental intrinsic semantics. For instance, spatially shifting an agent's coordinate features into an enemy's feature slot fundamentally alters the underlying Markov Decision Process (MDP) dynamics(~\cite{adaptaug2024, luo2024euclidean}). While data augmentation has recently demonstrated significant potential to enhance sample efficiency and representational robustness in RL and MARL(~\cite{yu2023esp}), the transformations must be carefully restricted to strictly preserve state-based semantic alignments. 

To generate robust positive pairs without violating these intrinsic state semantics, we designed a streamlined data augmentation pipeline tailored for 1D MARL state vectors. Our approach integrates two distinct, semantics-preserving operations:
\begin{itemize}[leftmargin=*, topsep=2pt, itemsep=2pt]
    \item \textbf{Random Masking / Cutout:} Elements within the state vector are randomly masked out (set to zero) with a probability $p_{mask}$. This acts as a localized information dropout mechanism, forcing the semantic embedder to reconstruct and align features even under severe partial observability or simulated sensor occlusion(~\cite{adaptaug2024}).
    \item \textbf{Gaussian Noise:} We continuously inject standard normal noise scaled by a factor of $\sigma$. This ubiquitous high-frequency noise simulates dynamic measurement perturbations, ensuring that the learned representations remain highly resilient to stochastic environmental dynamics(~\cite{yu2023esp}).
\end{itemize}

\begin{figure}[htbp]
    \centering
    \includegraphics[width=0.98\linewidth]{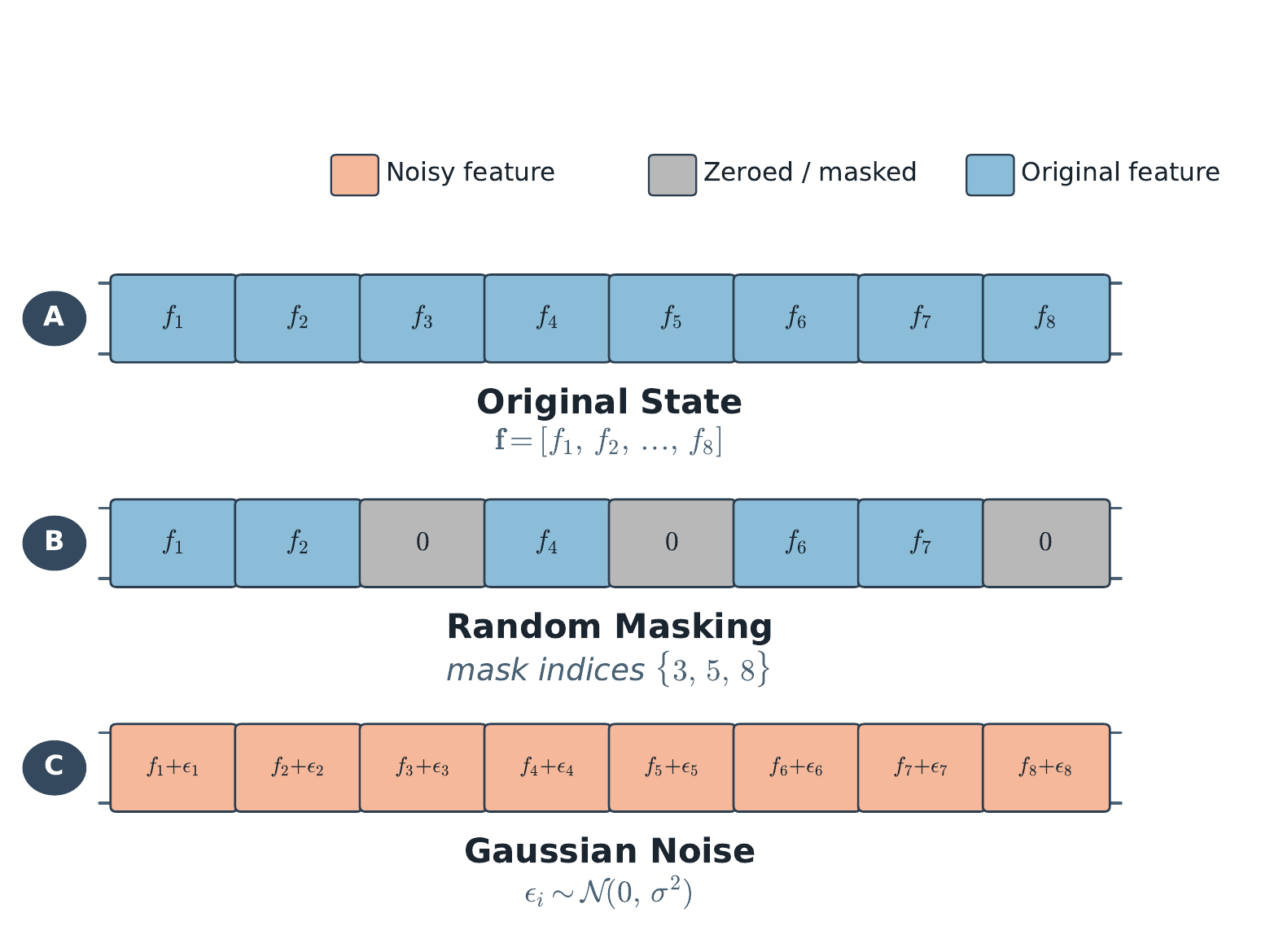}
    \vspace{-5pt}
    \caption{Conceptual illustration of the tailored data augmentation operations for 1D MARL state vectors. \textbf{(A)} The original state feature vector. \textbf{(B)} Random Masking drops out specific elements (e.g., indices 3, 5, 8) to simulate sensor occlusion without disrupting the alignment of other entities. \textbf{(C)} Gaussian Noise perturbs each individual feature with continuous noise ($\epsilon_i \sim \mathcal{N}(0, \sigma^2)$) to enhance measurement robustness.}
    \label{fig:aug_illustration}
\end{figure}

To provide a more intuitive understanding of our customized data augmentation pipeline for 1D state vectors, the specific operational formats are visually detailed in Figure \ref{fig:aug_illustration}. Unlike standard image transformations, our approach directly manipulates the numerical feature arrays to construct robust positive pairs while strictly preserving entity-level index alignment. Specifically, given an original state vector (Figure \ref{fig:aug_illustration}A), we employ \textit{Random Masking} to simulate information loss by zeroing out specific indices (Figure \ref{fig:aug_illustration}B), and \textit{Gaussian Noise} to perturb the continuous feature values (Figure \ref{fig:aug_illustration}C). By synergizing these operations, TCSE effectively learns representations that are robust to partial observability and invariant to environmental noise.

Furthermore, to systematically evaluate this combined data augmentation strategy, we conducted an extensive parametric study utilizing the \textbf{EMTC (QPLEX)} architecture as our fixed baseline. By keeping all core algorithmic components strictly unchanged and varying solely the data augmentation parameters, we detail the final test win-rates across various hyperparameter combinations on the \texttt{6h\_vs\_8z} and \texttt{corridor} maps in Tables \ref{tab:aug_6h8z} and \ref{tab:aug_corridor}. The empirical results consistently demonstrate that the framework achieves its peak performance at the optimal configuration of $p_{mask} = 0.2$ and $\sigma = 0.05$ across both scenarios. For instance, insufficient masking ($p_{mask} = 0.1$) limits the regularization effect, making the semantic embedder prone to overfitting local features, while excessive noise ($\sigma = 0.10$) heavily disrupts the inherent state structure. This rigorous evaluation verifies that our selected configuration provides the ideal balance between representation robustness and semantic feature retention in highly complex environments.

\begin{table}[htbp]
    \centering
    \caption{Parameter sensitivity of the combined augmentation strategy on the \texttt{6h\_vs\_8z} map. The optimal configuration is highlighted in bold.}
    \label{tab:aug_6h8z}
    \vspace{5pt}
    \begin{tabular}{cc|c}
        \toprule
        \textbf{Mask} ($p_{mask}$) & \textbf{Noise} ($\sigma$) & \textbf{Test Win} \\
        \midrule
        0.1 & 0.05 & 0.721 \\
        \textbf{0.2} & \textbf{0.05} & \textbf{0.748} \\
        0.4 & 0.05 & 0.742 \\
        0.2 & 0.10 & 0.701 \\
        \bottomrule
    \end{tabular}
\end{table}

\begin{table}[htbp]
    \centering
    \caption{Parameter sensitivity of the combined augmentation strategy on the \texttt{corridor} map. The table mirrors the experimental setup of \texttt{6h\_vs\_8z} to verify generalizability in a super-hard scenario.}
    \label{tab:aug_corridor}
    \vspace{5pt}
    \begin{tabular}{cc|c}
        \toprule
        \textbf{Mask} ($p_{mask}$) & \textbf{Noise} ($\sigma$) & \textbf{Test Win} \\
        \midrule
        0.1 & 0.05 & 0.922 \\
        \textbf{0.2} & \textbf{0.05} & \textbf{0.935} \\
        0.4 & 0.05 & 0.928 \\
        0.2 & 0.10 & 0.892 \\
        \bottomrule
    \end{tabular}
\end{table}

\subsection{Parametric Study on TCGM}
\label{Parametric Study on TCGM}
The temperature parameter $\tau$ in the Gaussian gating function (Eq. \eqref{eq:beta_gate}) dictates the strictness of the temporal consistency constraint. A larger $\tau$ produces a wider Gaussian curve, allowing episodic incentives (\cite{barto2012intrinsic, ng1999policy}) to pass through even when the temporal error $\Delta_{\text{mem}}$ is relatively large. Conversely, a smaller $\tau$ tightly restricts the distribution of incentives to only highly consistent trajectories.

\label{app:tau}
\begin{figure}[htbp]
    \centering
    \begin{subfigure}{0.48\textwidth}
        \centering
        \includegraphics[width=\linewidth]{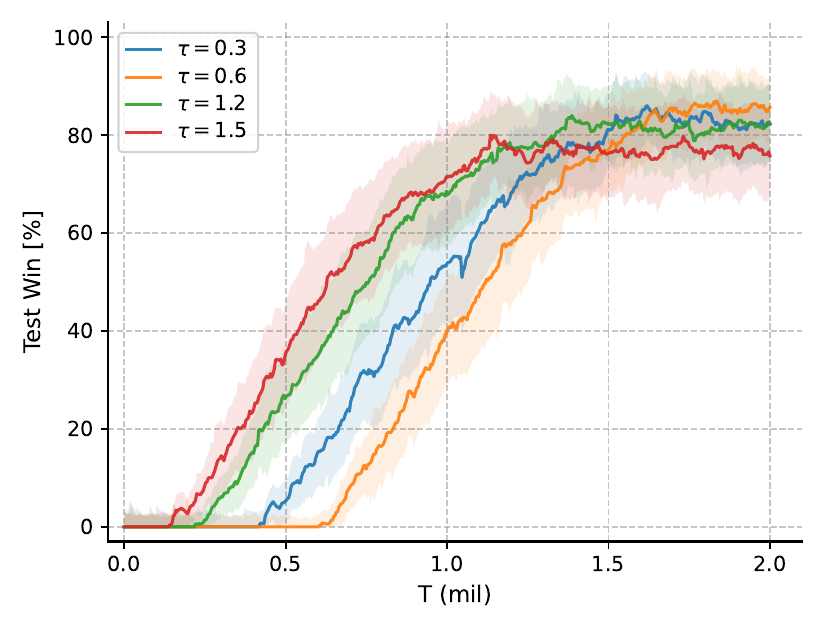}
        \caption{\texttt{3\_vs\_1WK} with \sysname{}}
    \end{subfigure}
    \hfill
    \begin{subfigure}{0.48\textwidth}
        \centering
        \includegraphics[width=\linewidth]{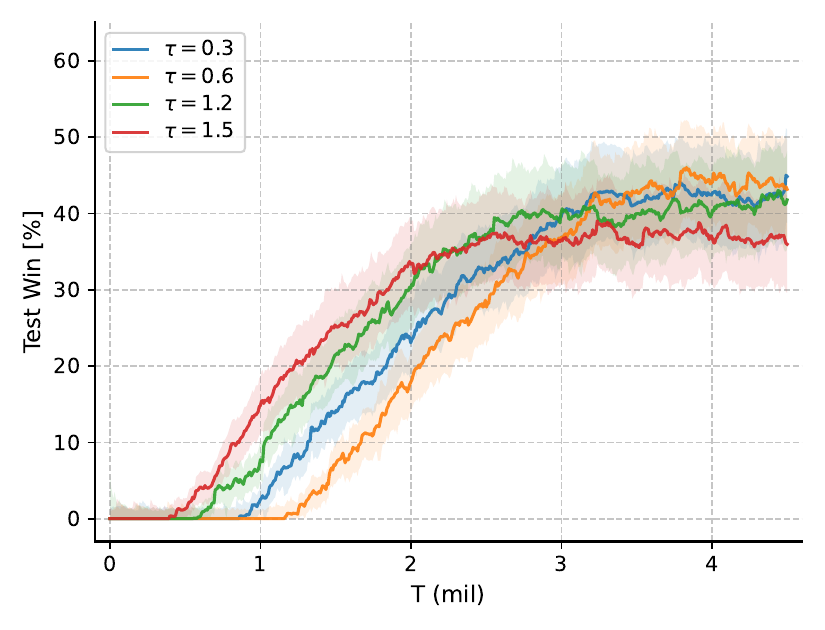}
        \caption{\texttt{CA\_hard} with \sysname{}(CDS)}
    \end{subfigure}
    \vspace{-5pt}
    \caption{Ablation study of the temporal consistency parameter $\tau$ using fixed values. The evaluation is conducted on two distinct scenarios: (a) \texttt{3\_vs\_1WK} using the \sysname{} architecture, and (b) the challenging \texttt{CA\_hard} using \sysname{}(CDS).}
    \label{fig:tau}
\end{figure}

We conduct an ablation study using fixed values of $\tau$ in $\{0.3, 0.6, 1.2, 1.5\}$. We evaluate these values across two distinct scenarios: the \texttt{3\_vs\_1WK} map using the standard \sysname{} architecture and the \texttt{CA\_hard} map using \sysname{}(CDS). Figure \ref{fig:tau} illustrates a consistent trend across both environments. An extremely large temperature like $\tau = 1.5$ achieves rapid performance growth early in training. A lenient gate allows many episodic rewards to pass. However, as noted in general reinforcement learning (\cite{sutton1998reinforcement}), without rigorous incentive constraints (\cite{van2016deep}), the unconstrained accumulation of auxiliary incentives can introduce significant estimation bias and systematic value overestimation (\cite{fujimoto2018addressing,henderson2018deep}). Consequently, this leniency eventually traps the policy in suboptimal behaviors and local optima (\cite{bellemare2016unifying}), where the agent prioritizes the maximization of unconstrained auxiliary incentives over the global task objective. Thus, the agent achieves a very low final win rate, making this value unsuitable for final adoption.

Conversely, strict temperatures rigorously filter out unreliable memories. They ensure robust model refinement and a higher performance plateau. Still, the initial learning speed is extremely slow. Notably, the experimental results show that $\tau = 0.6$ yields better overall performance than $\tau = 0.3$. This indicates that an overly strict gate blocks too many valid signals and harms the learning process. These contrasting observations clearly highlight the classic exploration and exploitation dilemma. 

To resolve this dilemma, we introduce a dynamic simulated annealing strategy. Based on the ablation results, we initialize the temperature at $1.2$ to encourage safe and rapid early exploration. We then gradually decay it to $0.6$ to ensure stable final convergence. The dynamic value of $\tau$ at training step $t$ follows a linear decay formulation:
\begin{equation}
    \tau(t) = \max \left( \tau_{end}, \tau_{start} - \frac{\tau_{start} - \tau_{end}}{T_{decay}} \cdot t \right)
\end{equation}
where we set $\tau_{start} = 1.2$ and $\tau_{end} = 0.6$. The parameter $T_{decay}$ represents the designated duration for the annealing process. Our final \sysname{} framework exclusively adopts this formula to regulate the temporal consistency gate throughout the entire training phase.

\subsection{Comparison of \sysname{} with MAPPO and DMIX on SMAC}
\label{app:vs_mappo}
In this subsection, we evaluate our framework against two prominent state of the art baselines. We select DMIX (\cite{sun2021dfac}) as a representative advanced value based method and we select MAPPO (\cite{yu2022surprising}) as a highly popular policy gradient method. We conduct these comparisons across three diverse SMAC maps. These maps include \texttt{3s\_vs\_5z}, \texttt{corridor}, and \texttt{6h\_vs\_8z}. The figures illustrate the experimental learning curves for our two variants, \sysname{}(QPLEX) and \sysname{}(CDS). Our method demonstrates a clear performance gap over both baselines.

\begin{figure}[htbp]
    \centering
    \begin{subfigure}{\textwidth}
        \centering
        \includegraphics[width=\linewidth]{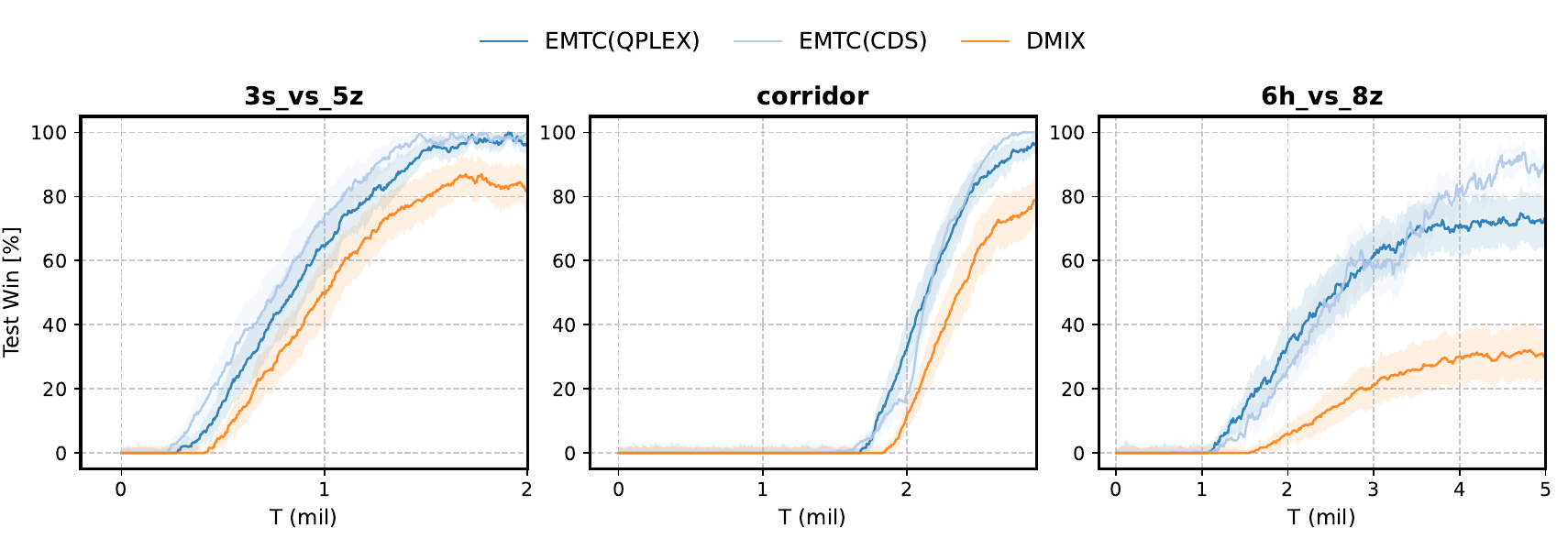}
        \caption{\sysname{} vs. DMIX}
        \label{fig:vs_dmix}
    \end{subfigure}
    
    \vspace{10pt} 
    
    \begin{subfigure}{\textwidth}
        \centering
        \includegraphics[width=\linewidth]{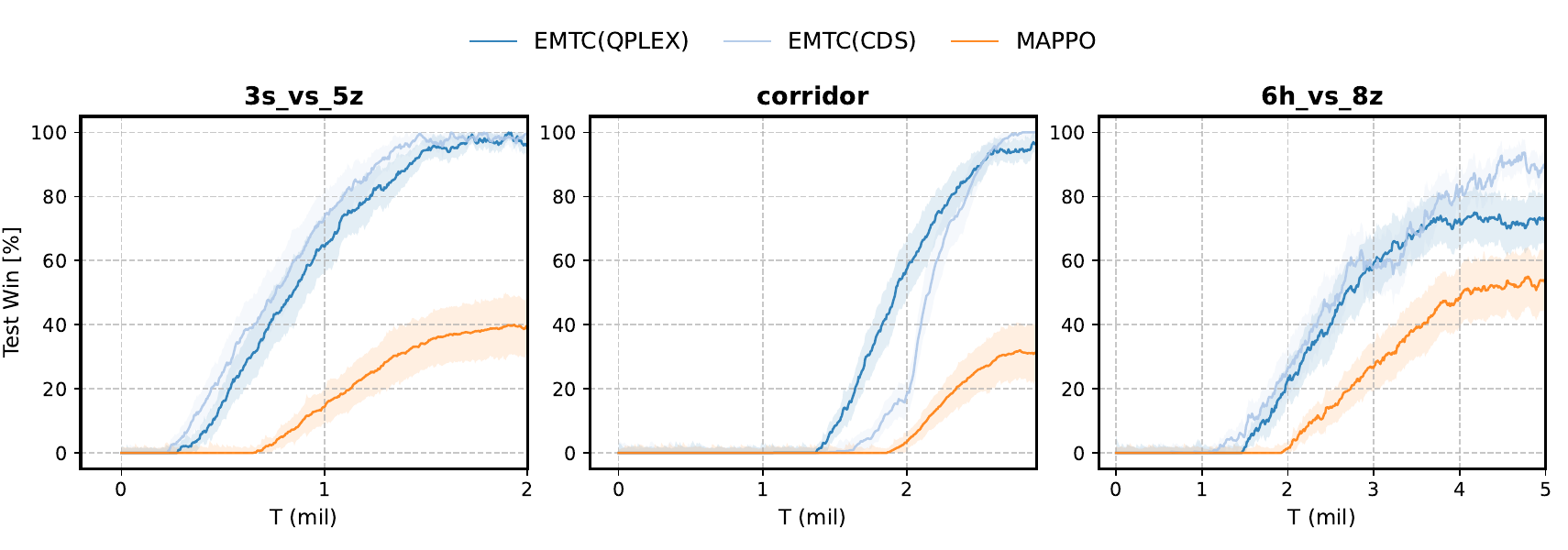}
        \caption{\sysname{} vs. MAPPO}
        \label{fig:vs_mappo}
    \end{subfigure}
    
    \caption{Performance comparison of \sysname{} variants against state of the art baselines. The evaluations span three diverse SMAC scenarios. The top row displays the comparison with DMIX, while The bottom row displays the comparison with MAPPO.}
    \label{fig:baseline_comparison}
\end{figure}

First, we compare our method with DMIX. DMIX uses distributional reinforcement learning to handle environmental randomness. However, it still relies on stepwise rewards. In super hard scenarios with severe reward sparsity, DMIX struggles to find successful joint actions. As Figure \ref{fig:vs_dmix} shows, DMIX learns slowly in \texttt{3s\_vs\_5z} and \texttt{corridor}. Furthermore, it suffers from a severe exploration deadlock in \texttt{6h\_vs\_8z}. It only achieves a very low final win rate. In contrast, our \sysname{} framework reuses historical high return trajectories. This memory mechanism directly guides agents toward rewarding states. It completely bypasses the exploration bottleneck and achieves superior performance.

Second, we compare our method with MAPPO. MAPPO is a highly popular on policy algorithm. The original paper (\cite{yu2022surprising}) notes that MAPPO can match off policy algorithms after extensive training. However, it exhibits significant limitations within our restricted training timesteps. We can see that \sysname{} performs better than MAPPO with an evident gap in Figure \ref{fig:vs_mappo}. Specifically, MAPPO suffers from severe sample inefficiency. The massive joint action space makes on policy exploration extremely inefficient here. MAPPO shows very slow progress and high performance variance across all three maps. Our \sysname{} framework completely avoids these issues. It maximizes sample efficiency through a robust semantic episodic buffer. As a result, our method drastically accelerates training and achieves a much higher final win rate.

\section{Algorithm}

\subsection{Training Algorithm for the TCSE Embedder}

\label{app:tcse}

Algorithm \ref{alg:tcse} outlines the optimization process for our semantic embedder. We periodically sample a dataset of historical transitions from the episodic buffer. Then, we divide this dataset into multiple mini batches. For each raw state, we apply spatial shift, random masking, and Gaussian noise. These operations generate a robust augmented view. Next, we feed both the raw state and its augmented view into the encoder network and this encoding step produces a positive pair of latent embeddings. 

The decoder network processes the raw state embedding and it reconstructs the original state and predicts the expected historical return. We compute the total objective loss using two main components. The first component is the contrastive loss. It aligns the latent embeddings of the positive pairs to extract robust semantic features. The second component is the autoencoder loss and it minimizes the reconstruction errors for both the state and the return. We use the weight parameter $\lambda_{rcon}$ to balance these two reconstruction targets, while using another parameter $\lambda_{cl}$ to scale the overall contrastive loss. 

We update the encoder and decoder networks via gradient descent and completely refresh the episodic buffer. Then, we use the updated encoder to project all historical raw states into a new latent space. This vital updating step effectively prevents stale representations and ensures highly accurate memory retrieval during the main policy training.

\begin{algorithm}[htbp]
\caption{Optimization for TCSE Embedder}
\label{alg:tcse}
\begin{algorithmic}[1]
\STATE \textbf{Parameter:} learning rate $\alpha$, number of training dataset $N$, batch size $B$, shift $s$, mask $p_{mask}$, noise $\sigma$, loss weights $\lambda_{rcon}, \lambda_{cl}$
\STATE Sample Training dataset $(s^{(i)}, H^{(i)}, t^{(i)})_{i=1}^N \sim \mathcal{D}_E$
\STATE Initialize weights $\phi, \psi \leftarrow \mathbf{0}$
\FOR{$i = 1$ \TO $\lfloor N/B \rfloor$}
    \STATE \textcolor{gray}{\# 1. Mini-batch fetching and Data Augmentation}
    \STATE Get a mini-batch of raw transitions $(s^{(j)}, H^{(j)}, t^{(j)})_{j=(i-1)B+1}^{iB}$
    \STATE Generate augmented states $s'^{(j)}$ via Shift ($s$), Mask ($p_{mask}$), and Noise ($\sigma$)
    
    \STATE \textcolor{gray}{\# 2. Latent Embedding and Contrastive Pairs}
    \STATE Compute latent embeddings for raw states $x^{(j)} = f_\phi(s^{(j)}|t^{(j)})$
    \STATE Compute latent embeddings for augmented states $x'^{(j)} = f_\phi(s'^{(j)}|t^{(j)})$
    
    \STATE \textcolor{gray}{\# 3. Decoder Reconstruction}
    \STATE Predict expected return $\bar{H}^{(j)} = f_\psi^H(x^{(j)}|t^{(j)})$
    \STATE Reconstruct original state $\bar{s}^{(j)} = f_\psi^s(x^{(j)}|t^{(j)})$
    
    \STATE \textcolor{gray}{\# 4. Explicit Loss Computation}
    \STATE Compute InfoNCE loss $\mathcal{L}_{cl}$ using positive pairs $(x^{(j)}, x'^{(j)})$
    \STATE Compute reconstruction loss $\mathcal{L}_{dCAE} = \frac{1}{B} \sum_j \left( \|H^{(j)} - \bar{H}^{(j)}\|_2^2 + \lambda_{rcon} \|s^{(j)} - \bar{s}^{(j)}\|_2^2 \right)$
    \STATE Compute total loss $\mathcal{L}(\phi, \psi) = \mathcal{L}_{dCAE} + \lambda_{cl}\mathcal{L}_{cl}$
    
    \STATE \textcolor{gray}{\# 5. Network Update}
    \STATE Update $\phi \leftarrow \phi - \alpha \frac{\partial \mathcal{L}}{\partial \phi}, \psi \leftarrow \psi - \alpha \frac{\partial \mathcal{L}}{\partial \psi}$
\ENDFOR
\STATE Re-project and update all memory embeddings $x \in \mathcal{D}_E$ with the updated $f_\phi$
\end{algorithmic}
\end{algorithm}

\subsection{Construction of the Episodic Memory Buffer}
\label{sec:appD2}
During centralized training, we continually evaluate the final episodic return. We check if the return reaches the maximum possible value $R_{max}$ or a specific success threshold $R_{thr}$. This success condition corresponds to defeating all enemies in SMAC or scoring a goal in GRF. When storing a successful trajectory $\mathcal{T}_\xi$ into the episodic buffer $\mathcal{D}_E$, we assign a positive desirability score. Specifically, we set $\xi(s) = 1$ for all states $s \in \mathcal{T}_\xi$.

To ensure efficient memory construction, we implement a desirability propagation mechanism. We propagate the positive desirability of a new state to any similar stored state within the distance threshold $\delta$. This explicit propagation provides a strong incentive for the agent to visit these similar states. Furthermore, stored memories generally remain in $\mathcal{D}_E$ until they become obsolete and get replaced. However, we actively update suboptimal records to maintain optimal buffer quality. The algorithm might detect a new desirable state near an existing suboptimal memory within the threshold $\delta$. In this case, we replace the old suboptimal observation with the new desirable one. This replacement essentially creates a memory shift toward highly rewarding states. Algorithm \ref{alg:memory} presents the complete memory construction procedure alongside these propagation and shift mechanisms.

\begin{algorithm}[htbp]
\caption{Episodic Memory Construction}
\label{alg:memory}
\begin{algorithmic}[1]
\STATE \textbf{Input:} Episodic trajectory $\mathcal{T} = \{s_0, a_0, r_0, ..., s_T\}$, Optimality $\xi_\mathcal{T}$, Episodic buffer $\mathcal{D}_E$
\STATE Initialize accumulated return $R_t = 0$
\STATE Initialize successor index $idx_{next} \leftarrow \text{Null}$
\FOR{$t = T$ \TO $0$}
    \STATE Compute latent embedding $x_t = f_\phi(s_t)$
    \STATE Update accumulated return $R_t \leftarrow r_t + \gamma R_t$
    \STATE Find the nearest neighbor $\hat{x}_t$ in $\mathcal{D}_E$ and retrieve its index $\hat{idx}_t$
    
    \IF{$\|x_t - \hat{x}_t\|_2 < \delta$}
        \STATE \textcolor{gray}{\# Update frequency and desirability counts}
        \STATE $N_{call}(\hat{x}_t) \leftarrow N_{call}(\hat{x}_t) + 1$
        \IF{$\xi_\mathcal{T} == 1$}
            \STATE $N_\xi(\hat{x}_t) \leftarrow N_\xi(\hat{x}_t) + 1$
        \ENDIF
        
        \STATE \textcolor{gray}{\# Update stored returns and representation}
        \IF{$\hat{\xi}_t == 0$ \AND $\xi_\mathcal{T} == 1$}
            \STATE $\hat{\xi}_t \leftarrow \xi_\mathcal{T}$ \quad \textcolor{gray}{\# Desirability propagation}
            \STATE $\hat{x}_t \leftarrow x_t, \hat{s}_t \leftarrow s_t$ \quad \textcolor{gray}{\# Memory shift}
            \STATE $\hat{H}_t \leftarrow R_t$
        \ELSIF{$\hat{H}_t < R_t$}
            \STATE $\hat{H}_t \leftarrow R_t$
        \ENDIF
        
        \STATE Update successor link: store $idx_{next}$ into the node at $\hat{idx}_t$
        \STATE $idx_{curr} \leftarrow \hat{idx}_t$
    \ELSE
        \STATE Add new memory tuple $\mathcal{D}_E \leftarrow (x_t, R_t, s_t, \xi_\mathcal{T}, idx_{next})$
        \STATE $idx_{curr} \leftarrow$ index of this newly added tuple
    \ENDIF
    
    \STATE \textcolor{gray}{\# Carry the current index to the previous timestep}
    \STATE $idx_{next} \leftarrow idx_{curr}$
\ENDFOR
\end{algorithmic}
\end{algorithm}

\subsection{Main Training Framework}
\label{app:main}
Algorithm \ref{alg:main} presents the main reinforcement learning loop of our framework. The agents continually interact with the environment to collect new trajectories. During this exploration process, the framework linearly decays the dynamic temperature parameter $\tau$. It then calls the memory construction subroutine to update the episodic buffer. 

Next, the algorithm evaluates the temporal consistency for the newly collected trajectory. For each transition $(s_t, a_t, r_t, s_{t+1})$, it retrieves the optimal historical return for the current state $s_t$ by finding its nearest neighbor in the episodic buffer $\mathcal{D}_E$. Crucially, to accurately capture the true suboptimality of the executed action, the algorithm also retrieves the expected optimal return for the successor state by querying $\mathcal{D}_E$ using the actual next state $s_{t+1}$ observed in the environment, rather than blindly following historical sequence links. The algorithm then calculates the exact temporal consistency error $\Delta_{\text{mem}}$. It uses the currently annealed temperature $\tau$ to compute the Gaussian gating coefficient $\beta(\Delta_{\text{mem}})$. The framework stores this coefficient alongside the regular transition data in the standard replay buffer. During the policy optimization phase, the algorithm samples a minibatch of transitions and calculates the standard unconstrained episodic reward. It strictly modulates this reward using the precomputed gating coefficient. The agent updates the joint action value networks using this gated loss function. Finally, the framework periodically triggers the embedder optimization subroutine to maintain semantic accuracy.

\begin{algorithm}[htbp]
\caption{Main Training Framework for EMTC}
\label{alg:main}
\begin{algorithmic}[1]
\STATE \textbf{Initialize:} Replay buffer $\mathcal{D}$, Episodic buffer $\mathcal{D}_E$
\STATE \textbf{Initialize:} Joint Q networks $\theta$, TCSE parameters $\phi, \psi$
\STATE \textbf{Initialize:} Annealing parameters $\tau_{start}$, $\tau_{end}$, $T_{decay}$
\WHILE{$t_{env} \leq t_{max}$}
    \STATE Interact with environment via $\epsilon$-greedy policy to collect trajectory $\mathcal{T}$
    
    \STATE \textcolor{gray}{\# 1. Anneal Temperature and Construct Memory}
    \STATE $\tau \leftarrow \max\left(\tau_{end}, \tau_{start} - \frac{\tau_{start} - \tau_{end}}{T_{decay}} \cdot t_{env}\right)$
    \STATE Run \textbf{Algorithm \ref{alg:memory}} to update $\mathcal{D}_E$ with trajectory $\mathcal{T}$
    
    \STATE \textcolor{gray}{\# 2. Evaluate Temporal Consistency Gating}
    \FOR{each transition $(s_t, a_t, r_t, s_{t+1})$ in $\mathcal{T}$}
        \STATE Retrieve optimal return $\mathcal{H}(s_t)$ via nearest neighbor search for $s_t$ in $\mathcal{D}_E$
        \STATE Retrieve successor optimal return $\mathcal{H}(s_{t+1})$ via nearest neighbor search for $s_{t+1}$ in $\mathcal{D}_E$
        \STATE Calculate temporal consistency error $\Delta_{\text{mem}} = |r_t + \gamma \mathcal{H}(s_{t+1}) - \mathcal{H}(s_t)|$
        \STATE Compute gating coefficient $\beta(\Delta_{\text{mem}}) = \exp\left(-\frac{\Delta_{\text{mem}}^2}{2\tau^2}\right)$
        \STATE Append transition tuple $(s_t, a_t, r_t, s_{t+1}, \beta(\Delta_{\text{mem}}))$ to replay buffer $\mathcal{D}$
    \ENDFOR
    
    \STATE \textcolor{gray}{\# 3. Main Policy Optimization}
    \FOR{$k = 1$ \TO $n_{circle}$}
        \STATE Sample a minibatch of $M$ transitions from $\mathcal{D}$
        \STATE Calculate the unconstrained episodic reward $r_p$
        \STATE Calculate the gated episodic reward $\tilde{r}_p = \beta(\Delta_{\text{mem}}) \cdot r_p$
        \STATE Update policy parameters $\theta$ using the gated TD loss $\mathcal{L}_\theta^\beta$
    \ENDFOR
    
    \STATE \textcolor{gray}{\# 4. Periodic TCSE Update}
    \IF{$t_{env} \bmod t_{emb} == 0$}
        \STATE Run \textbf{Algorithm \ref{alg:tcse}} to optimize parameters $\phi, \psi$
    \ENDIF
\ENDWHILE
\end{algorithmic}
\end{algorithm}

\clearpage
\newpage

\end{document}